# Enhancing Sample Efficiency and Exploration in Reinforcement Learning Through the Integration of Diffusion Models and Proximal Policy Optimization

**Tianci Gao[a,*], Konstantin A. Neusypin[a], Dmitry D. Dmitriev[a], Bo Yang[a], Shengren Rao[a]**

[a] Department IU-1 “Automatic Control Systems,” Bauman Moscow State Technical University, Moscow 105005, Russian Federation

[*] Corresponding author: Tianci Gao (email: gaotianci0088@gmail.com)

**Abstract**

Proximal Policy Optimization (PPO) is widely used in continuous control due to its robustness and stable training, yet it remains sample-inefficient in tasks with expensive interactions and high-dimensional action spaces. This paper proposes PPO-DAP (PPO with Diffusion Action Prior), a strictly on-policy framework that improves exploration quality and learning efficiency without modifying the PPO objective. PPO-DAP follows a two-stage protocol. Offline, we pretrain a conditional diffusion action prior on logged trajectories to cover the action distribution supported by the behavior policy. Online, PPO updates the actor–critic only using newly collected on-policy rollouts, while the diffusion prior is adapted around the on-policy state distribution via parameter-efficient tuning (Adapter/LoRA) over a small parameter subset. For each on-policy state, the prior generates multiple action proposals and concentrates them toward high-value regions using critic-based energy reweighting and in-denoising gradient guidance. These proposals affect the actor only through a low-weight imitation loss and an optional soft KL regularizer to the prior; importantly, PPO gradients are never backpropagated through offline logs or purely synthetic trajectories. We further analyze the method from a dual-proximal perspective and derive a one-step performance lower bound. Across eight MuJoCo continuous-control tasks under a unified online budget of 1.0M environment steps, PPO-DAP consistently improves early learning efficiency (area under the learning curve over the first 40 epochs, ALC@40) and matches or exceeds the strongest on-policy baselines in final return on 6/8 tasks, with modest overhead (1.18±0.04× wall-clock time and 1.05±0.02× peak GPU memory relative to PPO).

## 1. Introduction

On-policy reinforcement learning (RL) methods—typified by Proximal Policy Optimization (PPO)—remain one of the most widely adopted baselines for continuous control, largely because of their stable optimization behavior, simple training pipeline, and relative robustness to hyperparameter choices (Engstrom et al., 2020; Schulman et al., 2017). Nevertheless, their sample efficiency has been repeatedly criticized: when interaction is expensive or safety-critical, and the action space is high-dimensional, purely on-policy stochastic exploration can lead to slow early learning and premature convergence to suboptimal plateaus (Gan et al., 2024; Queeney et al., 2021).

Meanwhile, many real-world systems can provide logged trajectories collected by legacy controllers, simulators, or past experiments (Fu et al., 2020; Lee et al., 2022). This naturally raises the following question: **how can we exploit logged data to improve sample efficiency and exploration quality without violating PPO's strictly on-policy boundary**—i.e., policy gradients and advantage/ratio estimates must rely only on fresh rollouts from the current policy—and without introducing substantial online compute or engineering complexity?

Several research directions address parts of this problem but leave an important gap. Offline RL improves policies from static datasets via conservative objectives or behavior constraints, exemplified by CQL, TD3+BC, and IQL (Fujimoto & Gu, 2021; Kostrikov et al., 2021; Kumar et al., 2020), as well as model-based variants such as MOReL and MOPO (Kidambi et al., 2020; Yu et al., 2020). These methods can effectively exploit logs, but they typically depart from the on-policy update path and often require additional components (e.g., learned dynamics models), which increases deployment complexity.

In parallel, expressive generative models—including diffusion models—have been used for data augmentation, trajectory planning, and policy modeling. Recent work such as Diffuser and Diffusion Policy shows that diffusion models can provide powerful density modeling and control capabilities in action or trajectory space (Chi et al., 2023; Janner et al., 2022). However, many diffusion-RL approaches either replace the policy with a diffusion sampler, require extensive deployment-time fine-tuning, or mix offline log training with online updates in a way that blurs the boundary between on-policy and off-policy learning. More recent online diffusion-policy RL methods (e.g., QVPO, Q-score matching, SDAC, and diffusion sampling for energy-based policies) partially unify generation and optimization (Ding et al., 2024; Jain et al., 2024; Ma et al., 2025; Psenka et al., 2023), but they typically assume the diffusion network itself is the policy class, leading to a more complex training loop and higher sensitivity to value-estimation quality and numerical stability.

Overall, there remains no lightweight and implementation-friendly framework that (i) keeps PPO strictly on-policy, (ii) leverages logged data in a verifiable way, and (iii) benefits from the expressiveness of modern generative priors under a limited online interaction and compute budget.

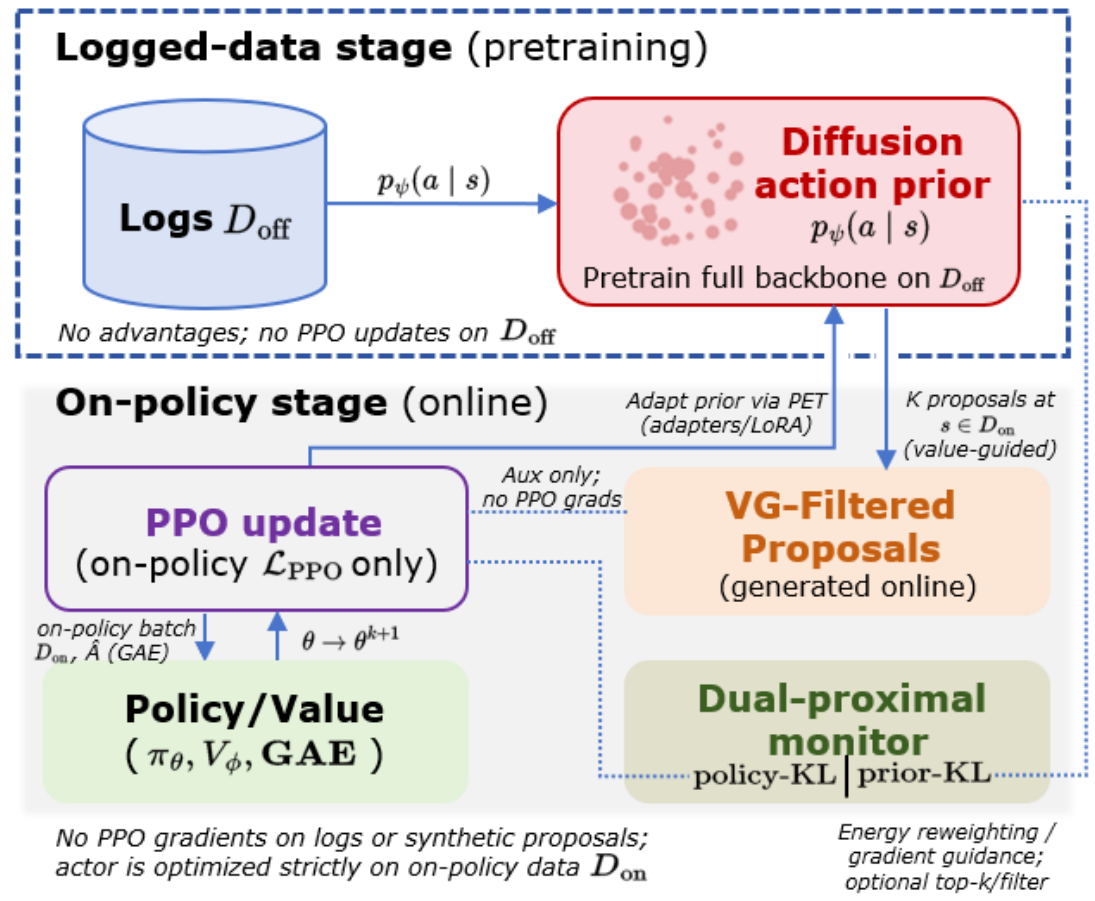

**Fig. 1. Overview of the two-stage PPO-DAP protocol.** *Logged-data stage:* a conditional diffusion action prior $p_\psi(a \mid s)$ is pretrained on offline logs $D_{\text{off}}$ only; no advantages are computed on $D_{\text{off}}$, and no PPO

updates are performed. *On-policy stage:* PPO updates the actor $\pi_\theta$ using standard PPO loss on newly collected rollouts $D_{\text{on}}$ only; the diffusion prior is adapted via parameter-efficient tuning (Adapter/LoRA) on a small parameter subset. For each on-policy state, the prior generates multiple value-guided action proposals (energy reweighting and optional in-denoising gradient guidance). Proposals affect the actor only through low-weight auxiliary imitation and optional soft KL-to-prior regularization; PPO gradients never backpropagate through offline logs or purely synthetic trajectories.

This paper adopts a complementary view: instead of treating diffusion models as policies or world models, we use them as an adaptable diffusion action prior that assists an on-policy PPO agent. We propose PPO-DAP (PPO with Diffusion Action Prior), a strictly on-policy framework with a two-stage protocol (Fig. 1). In the logged-data stage, we pretrain a conditional diffusion action prior to cover the behavioral action distribution supported by the offline logs; we never compute advantages on logged data and never perform PPO updates on offline trajectories. In the on-policy stage, PPO updates the policy only from newly collected on-policy rollouts, while the diffusion prior is adapted near the on-policy state distribution via parameter-efficient tuning (PET) that updates only small Adapter/LoRA modules. For each on-policy state, the prior generates multiple candidate actions; a lightweight critic provides Q estimates that guide candidates via energy-based reweighting and in-denoising gradient guidance. These candidates are used only in a small-weight auxiliary imitation loss and an optional soft prior-KL regularizer, and PPO gradients never pass through offline data or purely synthetic candidates.

From an optimization perspective, PPO-DAP exhibits a dual-proximal structure. PPO constrains the policy update via ratio clipping or a KL target, while PET constrains prior drift by restricting updates to low-rank adapters, which implicitly limits the prior KL between consecutive iterations. Under mild assumptions, this structure yields a per-update performance lower bound, where expected improvement is the standard on-policy PPO surrogate minus penalties associated with the policy KL, the prior KL, and value-guidance errors. Beyond the analysis, we also provide empirical evidence of strict on-policy behavior via dedicated on-policy audit metrics (e.g., OGLR, SPR, and PGShare) and by tracking policy-KL/prior-KL trajectories during training.

**Main contributions**

(1) **Diffusion action prior decoupled from the policy.** We introduce PPO-DAP, where a conditional diffusion model is used as an action prior that proposes candidates on on-policy states, while PPO remains the sole decision-making policy. Offline logs are used only to pretrain the prior and never enter PPO’s advantage estimation or policy-gradient pathway.

(2) **Value-guided proposals with lightweight regularization.** We design a value-guidance mechanism that combines energy-based reweighting and in-process gradient guidance to concentrate proposals in high-Q regions. Proposals influence the actor only through a small auxiliary imitation loss and an optional soft prior-KL term evaluated on the same on-policy states, leaving PPO’s main estimator unbiased and stable.

(3) **Parameter-efficient adaptation and a dual-proximal stability view.** We adapt the diffusion prior online via Adapter/LoRA-style parameter-efficient tuning, which keeps prior drift small and yields a dual-proximal structure (policy KL controlled by PPO; prior KL controlled by PET). We derive a corresponding per-update performance lower bound and empirically validate strict on-policy behavior and stability via audit metrics and KL trajectories.

(4) **Systematic evaluation under a unified interaction budget.** Under a fixed online budget of 1.0M environment steps across eight MuJoCo tasks, PPO-DAP improves early learning efficiency and matches or exceeds the strongest on-policy baselines on most tasks, while incurring only modest additional wall-clock time and GPU memory overhead. Ablations, t-SNE analyses, coverage scans, and compute-fairness studies further clarify the respective roles of the diffusion prior, value guidance, and PET and characterize practical trade-offs.

The remainder of this paper is organized as follows. Section 2 reviews related work on on-policy RL, offline and offline-to-online RL, generative models for control, diffusion guidance, and parameter-efficient adaptation. Section 3 formalizes the problem setting and the two-stage protocol. Section 4 details PPO-DAP, including the diffusion prior, value-guided proposals, PET updates, strict on-policy audits, and the dual-proximal analysis. Section 5 reports experiments on MuJoCo benchmarks, including ablations, compute fairness, coverage sensitivity, and value-guidance reliability, and discusses limitations and future directions.

## 2. Related work

Continuous-control RL methods often trade off training stability, sample efficiency, and practical deployability. This paper targets the intersection where (i) PPO-style updates remain strictly on-policy (policy gradients and advantage/ratio estimates are computed only from fresh rollouts), (ii) logged trajectories are exploited to improve early learning, and (iii) online compute overhead stays controllable (Engstrom et al., 2020; Levine et al., 2020; Schulman et al., 2017). Below we summarize the most relevant threads.

### 2.1 On-policy reinforcement learning and sample efficiency

Policy-gradient on-policy methods are widely used in continuous control, with PPO being a representative work (Schulman et al., 2017). By clipping the importance ratio or explicitly constraining the KL divergence, PPO limits each policy update to a proximal region, mitigating catastrophic policy collapse in high-dimensional control. Empirically, however, PPO performance is highly sensitive to implementation details and training "recipes" (Engstrom et al., 2020).

A line of research improves sample efficiency while retaining on-policy sampling by refining update boundaries and training schedules—for example, through improved clipping/boundary design (Cheng et al., 2022) or multi-phase training that increases value-learning reuse (Cobbe et al., 2021). Another line introduces stronger forms of sample/advantage reuse and off-policy corrections, such as PPO-ARC (Cheng et al., 2024) or explicit off-policy PPO frameworks (Meng et al., 2023). These methods can improve data reuse but often blur the strictly on-policy boundary and thus require additional clarification of bias sources and correction mechanisms.

In contrast, we improve sample efficiency via an explicit, data-driven action prior learned from logged trajectories, while keeping PPO's actor/critic updates strictly on-policy and auditable.

### 2.2 Offline RL and offline-to-online fine-tuning

Offline RL aims to learn policies from fixed datasets while addressing extrapolation errors induced by distribution shift (Fu et al., 2020; Levine et al., 2020). Representative methods enforce conservatism or behavioral constraints to keep policies within the data support: CQL penalizes out-of-distribution actions in the Q function (Kumar et al., 2020); TD3+BC augments TD3 with behavior cloning regularization (Fujimoto & Gu, 2021); and AWAC/IQL combine advantage-weighted behavior cloning or implicit value learning with conservative updates (Kostrikov et al., 2021; Nair et al., 2020). Another direction is model-based offline RL, which learns a dynamics model and performs planning or evaluation in the learned model (Kidambi et al., 2020; Yu et al., 2020).

In practice, a common paradigm is offline-to-online learning: policies are pre-trained using logged data (via BC or offline RL) and then fine-tuned with online interaction to reduce cold-start cost (Lee et al., 2022; Nakamoto et al., 2023; Nair et al., 2020). However, two tensions arise. First, offline objectives (often off-policy and conservative) are not aligned with PPO's on-policy objective. Second, some implementations continue to reuse offline data (or stale advantages) during fine-tuning, resulting in mixed on/off-policy updates whose on-policy boundary is difficult to define and verify (Cheng et al., 2024; Meng et al., 2023).

Our design follows an auditable boundary principle: offline logs are used only to pretrain an action prior; we never compute advantages on offline trajectories and never perform PPO updates using logged data. During online learning, actor/critic updates use only fresh on-policy rollouts, and offline data never re-enters the policy-gradient pathway.

### 2.3 Nearest-neighbor alignment and retrieval-based priors

An alternative to parametric generative priors is retrieval and nearest-neighbor alignment: given the current state (or an embedding), one retrieves nearby state–action pairs from a memory buffer and uses the retrieved actions as suggestions or soft constraints (e.g., local BC, local KL, or action-set constraints) (Humphreys et al., 2022; Jiang et al., 2023; Ran et al., 2023). These methods are simple, can provide fast cold-start, and can anchor exploration near the behavioral support when coverage is sufficient and embeddings are stable.

In high-dimensional continuous control, however, nearest-neighbor approaches have well-known limitations: retrieval quality can degrade rapidly under distribution shift or unstable representations (Humphreys et al., 2022); nonparametric neighborhoods struggle to represent multimodality and smooth interpolation in continuous action spaces; and most retrieval methods apply value-based ranking only after retrieval, making it harder to inject differentiable guidance during generation (Ran et al., 2023).

From our perspective, retrieval and PPO-DAP share the goal of anchoring exploration near the data support. PPO-DAP replaces nonparametric retrieval with a conditional diffusion prior that can represent multimodal continuous action distributions, supports smooth interpolation, and enables value-based reweighting and in-process gradient guidance, while PET mitigates online distribution drift.

### 2.4 Generative models for RL and policy modeling

Generative models have been used in RL both for data/candidate generation (as priors or proposal distributions) and as expressive policy or planning classes. In diffusion-based control, Diffuser frames control as conditional denoising in trajectory space (Janner et al., 2022), while Diffusion Policy generates action sequences via diffusion and has shown strong performance on robotic manipulation tasks (Chi et al., 2023). In offline RL, Diffusion-QL further couples diffusion policies with value functions to alleviate representational limitations of conventional policy classes (Wang et al., 2022).

Most diffusion-RL methods either treat the diffusion model as the policy (requiring optimization through the sampling process) or use diffusion as a world model/planner, which increases online training cost and engineering complexity. Moreover, training pipelines that mix offline and online data for policy optimization can blur the strictly on-policy boundary. In contrast, PPO-DAP uses diffusion only as an action prior/candidate generator, while the main policy remains a standard Gaussian PPO actor, making it easier to retain PPO's stability and implementation simplicity.

### 2.5 Diffusion guidance and online diffusion-policy RL

Guidance is a key reason diffusion models perform well across domains. Classifier-free guidance interpolates conditional and unconditional predictions to trade fidelity for diversity (Ho & Salimans, 2022). Energy/score-based guidance modifies the sampling objective or injects gradients during denoising to steer samples toward preferred regions (Du et al., 2023).

In control and RL, guidance signals are often derived from value functions or Q functions. Candidates can be reweighted by Q values, or Q can be treated as an energy function whose gradients are applied during denoising, pushing samples toward high-value regions. Recent frameworks directly optimize diffusion policies online, for instance via Q-weighted variational objectives (Ding et al., 2024), Q-score matching from reward/Q signals (Psenka et al., 2023), and efficiency-oriented online training variants (Ma et al., 2025). Although conceptually appealing, these approaches couple generation and optimization tightly, with three practical costs: (i) the diffusion network becomes the full policy class, leading to large models and more complex training; (ii) optimization through diffusion sampling can be high-variance and numerically sensitive; and (iii) mixed offline/online updates can further complicate strict on-policy verification.

PPO-DAP instead decouples diffusion from policy optimization: the diffusion prior generates candidates only on current on-policy states and uses energy reweighting and in-process gradient guidance to improve candidate quality, while PPO remains strictly on-policy and candidates influence the actor only through small-weight auxiliary terms.

### 2.6 Parameter-efficient fine-tuning and diffusion adaptation

Parameter-efficient fine-tuning (PEFT) adapts large models by updating only a small number of additional parameters (e.g., adapters or low-rank matrices), substantially reducing compute and storage cost (Houlsby et al., 2019; Hu et al., 2021). For diffusion models, PEFT has also been shown to enable effective adaptation while keeping the backbone frozen (Xie et al., 2023).

In control settings, full fine-tuning of a large diffusion model during online learning can be costly and unstable. PPO-DAP uses PEFT for online prior adaptation: the diffusion backbone is pretrained offline for expressiveness, then frozen online while small Adapter/LoRA modules are updated to track distribution drift around on-policy states. Together with PPO's proximal constraint on the policy KL, this forms the dual-proximal view used to explain PPO-DAP's stability and efficiency.

### 2.7 Summary and comparison

To summarize, prior work covers different parts of our target setting. PPO and on-policy variants offer stable improvement but remain sample-inefficient (Engstrom et al., 2020; Schulman et al., 2017). Offline RL and offline-to-online pipelines can exploit logs but typically rely on off-policy objectives or mixed updates that require extra care to define and audit a strict on-policy boundary (Levine et al., 2020; Lee et al., 2022; Nakamoto et al., 2023). Diffusion-based policies offer strong expressiveness and guidance but often incur higher online cost and more complex training (Chi et al., 2023; Ding et al., 2024; Psenka et al., 2023). Retrieval-based methods can anchor exploration near the data support but are sensitive to representation and coverage and provide limited in-process guidance (Humphreys et al., 2022; Ran et al., 2023). Finally, PEFT provides an economical mechanism for online adaptation but must be coordinated with RL update boundaries (Houlsby et al., 2019; Hu et al., 2021; Xie et al., 2023). Representative approaches are compared under these desiderata in **Table 1**.

Table 1. Comparison of representative approaches under three desiderata: strictly on-policy, logged-data utilization, and controllable compute

| Approach | Strictly on-policy gradients | Logged data in policy gradients / advantages | Explicit action support / prior | Value-guided high-Q proposals | Additional online compute |
|---|---|---|---|---|---|
| PPO / typical on-policy improvements (Schulman et al., 2017; Engstrom et al., 2020; Cheng et al., 2022; Cobbe et al., 2021) | ✓ | ✗ | ✗ | △ (via exploration noise and critic) | Low |
| Offline RL (CQL / TD3+BC / IQL) (Kumar et al., 2020; Fujimoto and Gu, 2021; Kostrikov et al., 2021; Nair et al., 2020) | ✗ | ✓ | △ (primarily via conservative / behavior-cloning constraints) | △ / ✓ (method-dependent) | Medium |
| Offline-to-online (pretrain + online RL) (Nair et al., 2020; Lee et al., 2022; Nakamoto et al., 2023) | △ (implementation-dependent) | △ | △ | △ | Medium |
| Nearest-neighbor / retrieval priors (Humphreys et al., 2022; Ran et al., 2023; Jiang et al., 2023) | ✓ (if no offline gradient mixing) | ✗ | ✓ (nonparametric local support) | △ (often via post-retrieval ranking) | Medium (retrieval / indexing) |
| Diffusion-as-policy (offline / online) (Janner et al., 2022; Wang et al., 2022; Chi et al., 2023; Ding et al., 2024; Psenka et al., 2023) | △ / ✗ (training-dependent) | △ (can be mixed) | ✓ | ✓ | High |
| Ours: PPO-DAP (diffusion prior + VG + PEFT) | ✓ | ✗ | ✓ (parametric conditional prior; adaptable) | ✓ (energy reweighting and in-process gradient guidance) | Medium (controllable) |

**Notes.** ✓: desideratum clearly satisfied; ✗: not satisfied; △: partially satisfied or implementation-dependent. "Additional online compute" is reported qualitatively (Low / Medium / High) relative to a standard PPO baseline.

## 3. Preliminaries and Problem Formulation

This section reviews the Markov decision process (MDP) formulation, value functions, and policy-gradient methods, and then formalizes the two-stage protocol of logged-data pretraining followed by strictly on-policy fine-tuning, which underpins our algorithm design and theoretical analysis.

### 3.1 Reinforcement learning and Markov decision processes

We model the environment as a discounted MDP

$$\mathcal{M} = \langle \mathcal{S}, \mathcal{A}, P, R, \gamma \rangle,$$

where $S$ and $A$ denote the state and action spaces, respectively; $\mathrm{P}(s' \mid s, a)$ is the transition kernel; $\mathrm{R}(s, a)$ is the immediate reward; and $\gamma \in [0,1)$ is the discount factor (Puterman, 1994; Sutton & Barto, 2018). A stochastic policy $\pi_\theta(a \mid s)$ maps each state to a distribution over actions, parameterized by $\theta$. At time step $t$, the agent observes $s_t$, samples $a_t \sim \pi_\theta(\cdot \mid s_t)$, receives $r_t = R(s_t, a_t)$, and transitions to $s_{t+1} \sim P(\cdot \mid s_t, a_t)$.

The objective is to maximize the expected discounted return:

$$J(\theta) = \mathbb{E}_{\tau \sim (\pi_\theta, P)} \left[ \sum_{t=0}^{\infty} \gamma^t r_t \right], \tag{1}$$

where $\tau = (s_0, a_0, r_0, s_1, \dots)$ denotes a trajectory sampled from the rollout distribution induced jointly by $\pi_\theta$ and the environment dynamics $P$.

**Data notation.** We use three types of data throughout this paper:

- **Logged dataset** $D_{\text{off}}$: a static offline dataset consisting of transitions $(s, a, r, s')$, collected by an unknown or legacy behavior policy $\mu$ and fixed during training (Fu et al., 2020; Levine et al., 2020).
- **On-policy batch** $D_{\text{on}}^{(k)}$: the batch of transitions collected at the $k$-th online iteration using the current policy: $D_{\text{on}}^{(k)} = \{(s_t, a_t, r_t, s_{t+1})\}_{t=1}^{T_k}$.
- **Synthetic proposal set** $D_{\text{syn}}^{(k)}$: a set of candidate actions generated by an action prior on **the same on-policy states**: $D_{\text{syn}}^{(k)} = \left\{ \left(s_t, \tilde{a}_{t,j}\right) : s_t \in D_{\text{on}}^{(k)}, \tilde{a}_{t,j} \sim p_\psi(\cdot \mid s_t), j = 1, \dots, K \right\}$.

Importantly, $D_{\text{syn}}^{(k)}$ shares the state marginal with $D_{\text{on}}^{(k)}$, while actions are sampled from the learned prior. We will explicitly enforce that these proposals do not enter PPO's importance ratio or advantage estimation (Section 3.3 and Section 4.5).

### 3.2 Value functions and advantage estimation

Given a policy $\pi$, the state-value function $V^\pi$ and action-value function $Q^\pi$ are defined as:

$$V^\pi(s) = \mathbb{E}_\pi \left[ \sum_{t=0}^{\infty} \gamma^t R(s_t, a_t) \mid s_0 = s \right], \tag{2}$$

$$Q^\pi(s, a) = \mathbb{E}_\pi \left[ \sum_{t=0}^{\infty} \gamma^t R(s_t, a_t) \mid s_0 = s, a_0 = a \right], \tag{3}$$

The advantage function measures the relative desirability of an action compared to the state baseline:

$$A^{\pi}(s,a) = Q^{\pi}(s,a) - V^{\pi}(s).$$

In practice, we adopt Generalized Advantage Estimation (GAE) to compute a low-variance estimator $\hat{A}_t$ (Schulman et al., 2016), which will be used to build the PPO surrogate objective.

### 3.3 Policy gradients and the PPO surrogate objective

Given the RL objective $J(\theta)$ defined in Section 3.1, policy-gradient methods optimize $\theta$ using gradient estimates derived from on-policy trajectories (Sutton et al., 2000; Schulman et al., 2016). Proximal Policy Optimization (PPO) is a widely used on-policy algorithm that stabilizes these updates by restricting the deviation between successive policies (Schulman et al., 2017).

Let $\theta_{\text{old}}$ denote the parameters of the behavior policy that generated the current on-policy batch. PPO defines the importance ratio

$$r_t(\theta) = \frac{\pi_\theta(a_t \mid s_t)}{\pi_{\theta_{\text{old}}}(a_t \mid s_t)}, \tag{4}$$

and maximizes the clipped surrogate objective

$$\mathcal{L}_{\text{PPO}}(\theta) = \mathbb{E}\Big[min\Big(r_t(\theta)\hat{A}_t, \text{clip}(r_t(\theta), 1-\epsilon, 1+\epsilon)\hat{A}_t\Big)\Big], \tag{5}$$

where $\epsilon > 0$ controls the effective update magnitude. The value function $V_\phi(s)$ is trained via TD/GAE regression to produce low-variance advantage estimates $\hat{A}_t$ (Schulman et al., 2016), which are then used in Eq. (5) as the main learning signal for the actor.

### 3.4 Problem setup and the two-stage protocol

In many real-world scenarios, online interactions are costly or safety-critical, while one may have access to historical logged trajectories collected by legacy controllers, simulators, or past experiments. Directly mixing offline data into on-policy updates can introduce distribution shift and biased gradient estimates, thereby blurring PPO's "fresh-rollout-only" boundary (Levine et al., 2020). We therefore adopt a strictly separated two-stage protocol that leverages logged data without compromising strict on-policy learning.

**Stage I: Logged-data pretraining**

From the static dataset $D_{\text{off}}$, we learn a conditional diffusion-based **action prior** $p_\psi(a \mid s)$, parameterized by a conditional diffusion model (Ho et al., 2020). This stage performs supervised / generative training only, and may optionally warm-start the policy and value networks with supervised losses. Crucially, it satisfies the following constraints:

- **No advantage estimation** $\hat{A}_t$ is computed on $D_{\text{off}}$;
- **No PPO updates** (i.e., no use of Eqs. (4)-(5) or the corresponding policy gradient) are performed on $D_{\text{off}}$;
- No environment dynamics model is learned; the focus is on learning $p_\psi(a \mid s)$ (and optionally supervised value initialization).

**Stage II: Online strictly on-policy fine-tuning**

At online iteration $k$, we collect a fresh on-policy batch $D_{\text{on}}^{(k)}$ with the current policy $\pi_\theta$. The actor is updated **only** on $D_{\text{on}}^{(k)}$ using Eq. (5), and the critic is updated using TD/GAE on the same on-policy batch. Meanwhile, the action prior $p_\psi$ is adapted in a parameter-efficient manner—updating only a small subset $\psi_{\text{PET}} \subset \psi$ (e.g., Adapter/LoRA parameters) near the on-policy state distribution (Houlsby et al., 2019; Hu et al., 2021). For each on-policy state $s_t \in D_{\text{on}}^{(k)}$, the adapted prior generates candidate actions to form $D_{\text{syn}}^{(k)}$.

The synthetic proposals are used only for:

- a lightweight auxiliary imitation loss (auxiliary BC), and
- an optional soft prior KL regularizer ("soft anchoring" toward the prior).

We emphasize the following strict on-policy boundary, which will be enforced in implementation (Section 4.5):

- No PPO gradient is backpropagated through $D_{\text{off}}$;
- $D_{\text{syn}}^{(k)}$ is excluded from PPO's importance ratio $r_t(\theta)$ and advantage estimation $\hat{A}_t$; Eqs. (4)-(5) are determined solely by $D_{\text{on}}^{(k)}$;
- The diffusion prior does not replace the policy and does not model environment dynamics (unlike model-based offline RL).

This protocol enables a clear division of labor: PPO provides stable on-policy policy improvement, while the diffusion model acts as an adaptive action prior that (i) anchors exploration within a data-supported region, and (ii) supplies value-aware proposals to accelerate learning.

# 4. Algorithm Design (PPO-DAP)

This section presents the full design of PPO-DAP (PPO with Diffusion Action Prior). The key idea is to decouple decision making (a standard Gaussian PPO actor–critic) from generation (a diffusion action prior), and to couple them only via small-weight regularizers evaluated on the same on-policy states with explicit gradient isolation. This design aims to improve early exploration and sample efficiency using logged-data priors without violating PPO's strict on-policy boundary (Schulman et al., 2017; Engstrom et al., 2020).

## 4.1 Two-stage decoupled architecture and design goals

As illustrated in Fig. 2, PPO-DAP separates:

- a **decision module**: a Gaussian PPO actor $\pi_\theta$ and a critic (value head $V_\phi$, plus a lightweight $Q$ −head $\hat{Q}_\phi$ for proposal guidance), and
- a **generation module**: a conditional diffusion action prior $p_\psi(a \mid s)$.

The two modules interact **only** through lightweight regularizers on the same on-policy states, with stop-gradient barriers preventing PPO gradients from flowing into the prior.

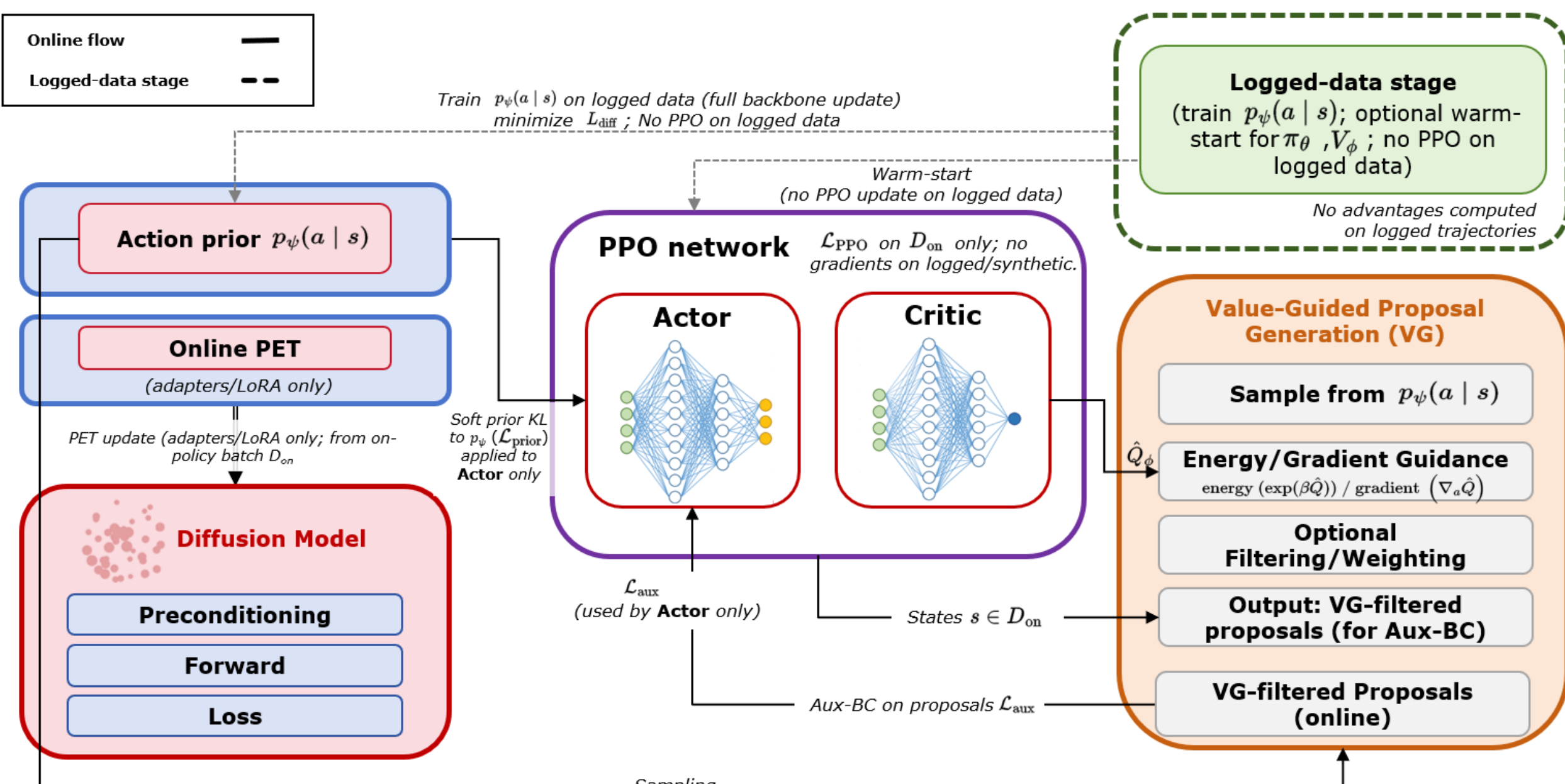

**Fig. 2.** Overall architecture of PPO-DAP. *In the logged-data stage (dashed arrows), a conditional diffusion model* $p_\psi(a \mid s)$ *is trained as an action prior and can optionally warm-start the actor–critic, while no advantages are computed and no PPO updates are performed on logged trajectories. In the online stage (solid arrows), PPO is updated only on fresh on-policy rollouts* $D_{on}$ *; the prior is adapted via parameter-efficient tuning (Adapters/LoRA) and generates value-guided candidate actions (energy/gradient guidance with optional filtering), which enter the actor only through small prior-KL and auxiliary imitation losses. Setting the adaptation frequency* $f = 0$ *corresponds to the No-PET configuration.*

**Logged-data stage (pretraining).** We train the conditional diffusion prior $p_\psi(a \mid s)$ on $D_{\text{off}}$ with all backbone parameters unfrozen. Optionally, we warm-start $\pi_\theta$ and $V_\phi$ via supervised learning. During this stage:

- no advantages are computed on logged trajectories,
- no PPO updates are performed,
- no KL-based coupling between policy and prior is imposed.

**Online stage (strictly on-policy).** We collect fresh rollouts $D_{\text{on}}^{(k)}$ from $\pi_\theta$. PPO updates the actor using only $D_{\text{on}}^{(k)}$. The diffusion prior is adapted near the on-policy state distribution via PET (updating only $\psi_{\text{PET}}$, e.g., Adapters/LoRA), and generates value-guided candidate actions on the same states. These candidates enter only auxiliary / regularization branches with small weights, without altering PPO's sampling distribution or importance ratios.

**Why decouple?** If the diffusion model is used as the policy itself, online optimization typically requires differentiating through the sampling procedure or adopting bespoke variational/score-matching objectives, which increases gradient variance and engineering complexity (Chi et al., 2023; Psenka et al., 2023; Ding et al., 2024). In contrast, PPO-DAP keeps the PPO actor–critic update unchanged and treats diffusion as a plug-in proposal module with explicit gradient blocking; therefore it can be integrated into existing PPO codebases with minimal modifications (no changes to advantage estimation, importance ratios, or the PPO optimizer), while the added overhead is controlled by a small set of transparent knobs $(K, N_{\text{steps}}, f, r)$. PPO-DAP thus follows a lighter route: PPO remains the sole decision-maker, while the diffusion model serves as an adaptive proposal generator and a soft regularizer, retaining PPO's robust "recipe" and implementation simplicity (Schulman et al., 2017; Engstrom et al., 2020).

### 4.2 Diffusion action prior: logged-data training and online PET

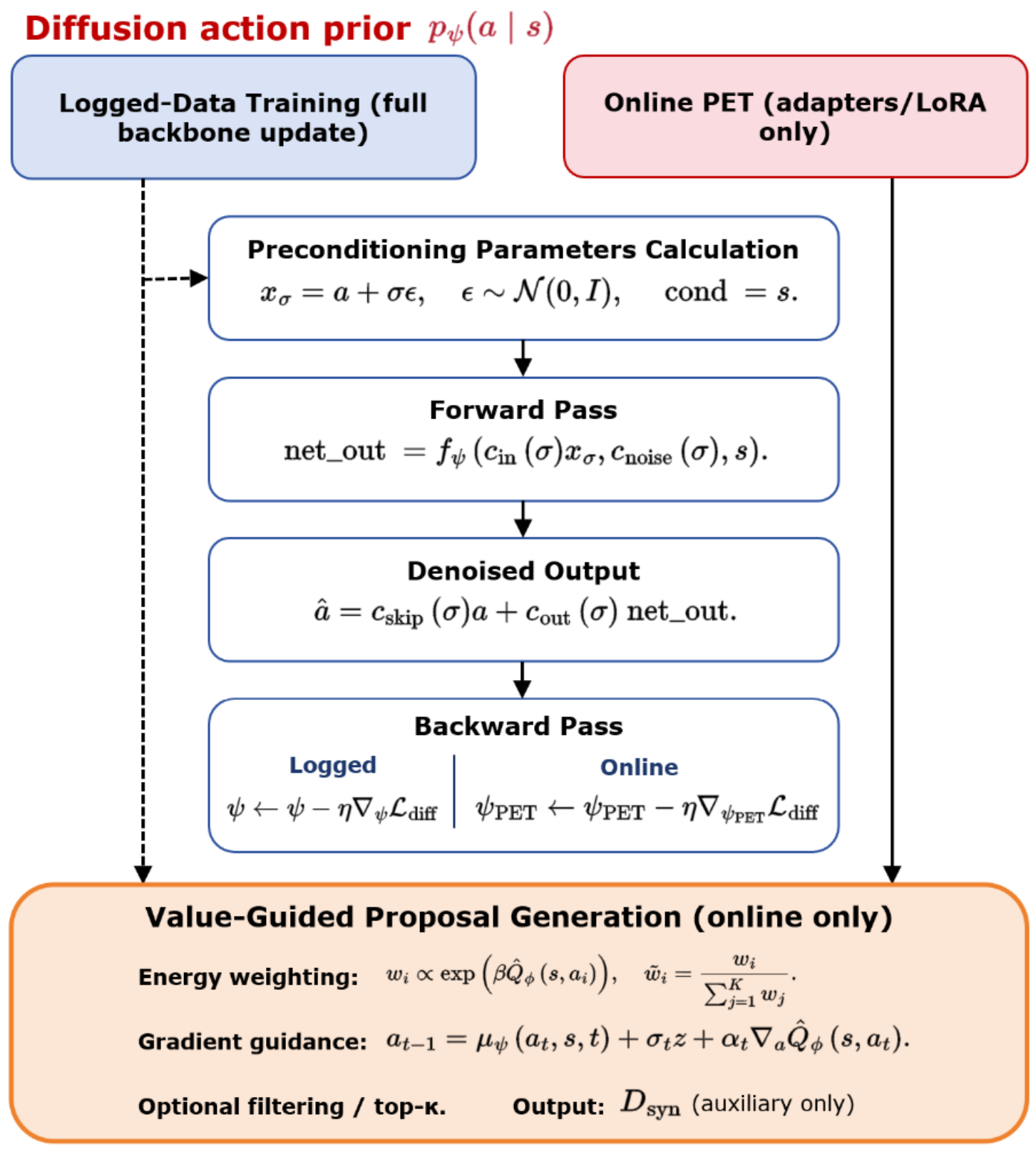

**Fig. 3.** Diffusion action prior with logged-data training, online PET adaptation, and value-guided proposal generation. *The logged phase trains the full diffusion backbone with a denoising loss, while the online PET phase updates only adapters/LoRA and uses value guidance (energy weighting or gradient guidance) to produce filtered proposal actions* $D_{syn}$ *used in auxiliary losses.*

We treat the diffusion prior as a conditional distribution $p_\psi(a \mid s)$ defined implicitly by a denoising diffusion process. The corresponding pretraining and online adaptation pipeline is illustrated in Fig. 3. Given $(s, a) \sim D_{\text{off}}$, we follow standard diffusion training to corrupt the action with noise and regress the clean action (Ho et al., 2020; Nichol & Dhariwal, 2021; Karras et al., 2022). Let

$$x_\sigma = a + \sigma\epsilon, \epsilon \sim \mathcal{N}(0, I), \sigma \sim \mathcal{P}(\sigma),$$

where $\mathcal{P}(\sigma)$ is the noise schedule. Let $\hat{a}_\psi(s, x_\sigma, \sigma)$denote the denoiser output (e.g., "predicting the clean action" under a chosen parameterization). The denoising loss is:

$$\mathcal{L}_{\text{diff}}(\psi) = \mathbb{E}_{(s,a)\sim D_{\text{off}},\sigma,\epsilon}\left[\left\|\hat{a}_{\psi}(s, x_{\sigma}, \sigma) - a\right\|_2^2\right]. \quad (6)$$

**Online PET (parameter-efficient tuning).** During online learning, we freeze the diffusion backbone and update only a small parameter subset $\psi_{\text{PET}} \subset \psi$ (e.g., Adapters/LoRA). We keep the same denoising objective but replace the data distribution by $D_{\text{on}}^{(k)}$:

$$\mathcal{L}_{\text{PET}}(\psi_{\text{PET}}) = \mathbb{E}_{(s,a)\sim D_{\text{on}}^{(k)},\sigma,\epsilon}\left[\left\|D_{\psi_{\text{PET}}}(s, a + \sigma\epsilon; \sigma) - a\right\|_2^2\right].$$

The motivation is practical: frequent full fine-tuning of a large diffusion model during online RL is often costly and unstable, while Adapters/LoRA allow small, stable updates with reduced compute and memory footprints (Houlsby et al., 2019; Hu et al., 2021). Parameter-efficient adaptation has also been validated for diffusion model transfer (Xie et al., 2023). With low rank $r$ and moderate update frequency $f$, the extra online overhead scales roughly with $\mathcal{O}(f \cdot r)$.

### 4.3 Value-guided proposals (VG)

At online iteration $k$, for each on-policy state $s \in D_{\text{on}}^{(k)}$, we sample $K$ candidate actions $\{a_i\}_{i=1}^{K} \sim p_{\psi}(\cdot \mid s)$ and guide them using a lightweight action-value head $\hat{Q}_{\phi}(s, a)$. The resulting proposals are used only for auxiliary/regularization losses, and **never** enter PPO's importance ratios. We consider two complementary guidance mechanisms.

**(i) Energy-based reweighting (post hoc guidance)**

We assign weights via a softmax over estimated action values:

$$w_i \propto \exp\left(\beta \hat{Q}_{\phi}(s, a_i)\right), \qquad \sum_{i=1}^{K} w_i = 1, \quad (7)$$

We then resample proposals according to $\{w_i\}$ to form $D_{\text{syn}}^{(k)}$. This resembles the maximum-entropy / energy-based view of sampling actions in proportion to an exponentiated preference signal (Haarnoja et al., 2018; Peng et al., 2019), and aligns with Q-weighted policy optimization intuitions (Ding et al., 2024). To mitigate mode collapse when the critic is unreliable early in training, we linearly anneal $\beta$ from 0 to 1 over the first 30% of iterations.

**(ii) In-process gradient guidance (during denoising)**

During diffusion sampling, at denoising step $t$, we shift the intermediate sample $a_t$ along the action-gradient of $\hat{Q}_{\phi}$:

$$a_{t-1} = \mu_{\psi}(a_t, s, t) + \sigma_t z + \alpha_t \nabla_a \hat{Q}_{\phi}(s, a_t), \quad (8)$$

where $\mu_{\psi}(\cdot)$ and $\sigma_t$ are given by the denoiser and the noise schedule, respectively, and $z \sim \mathcal{N}(0, I)$. This follows the general idea of gradient/energy guidance in diffusion models, where a differentiable preference function steers sampling toward desirable regions while preserving diversity (Dhariwal & Nichol, 2021; Ho & Salimans, 2022), and echoes value-guided diffusion planning ideas (Janner

et al., 2022).

We use a noise-aware schedule:

$$\alpha_t = \alpha_{max}\left(1 - \frac{\sigma_t}{\sigma_{max}}\right), \alpha_{max} = 0.3,$$

and clip the guidance term for numerical stability:

$$\left\|\alpha_t \nabla_a \hat{Q}_\phi(s, a_t)\right\|_2 \leq c, c = 0.1.$$

Intuitively, early high-noise steps preserve exploration, while later low-noise steps apply stronger "value sharpening."

**Lightweight usage.** To maintain the role of "auxiliary regularization," we cap the proportion of proposal samples in each actor update batch to at most $\approx 20\%$ and optionally apply top-$k$ filtering before constructing $D_{\text{syn}}^{(k)}$.

### 4.4 Policy update: PPO with small-weight regularization

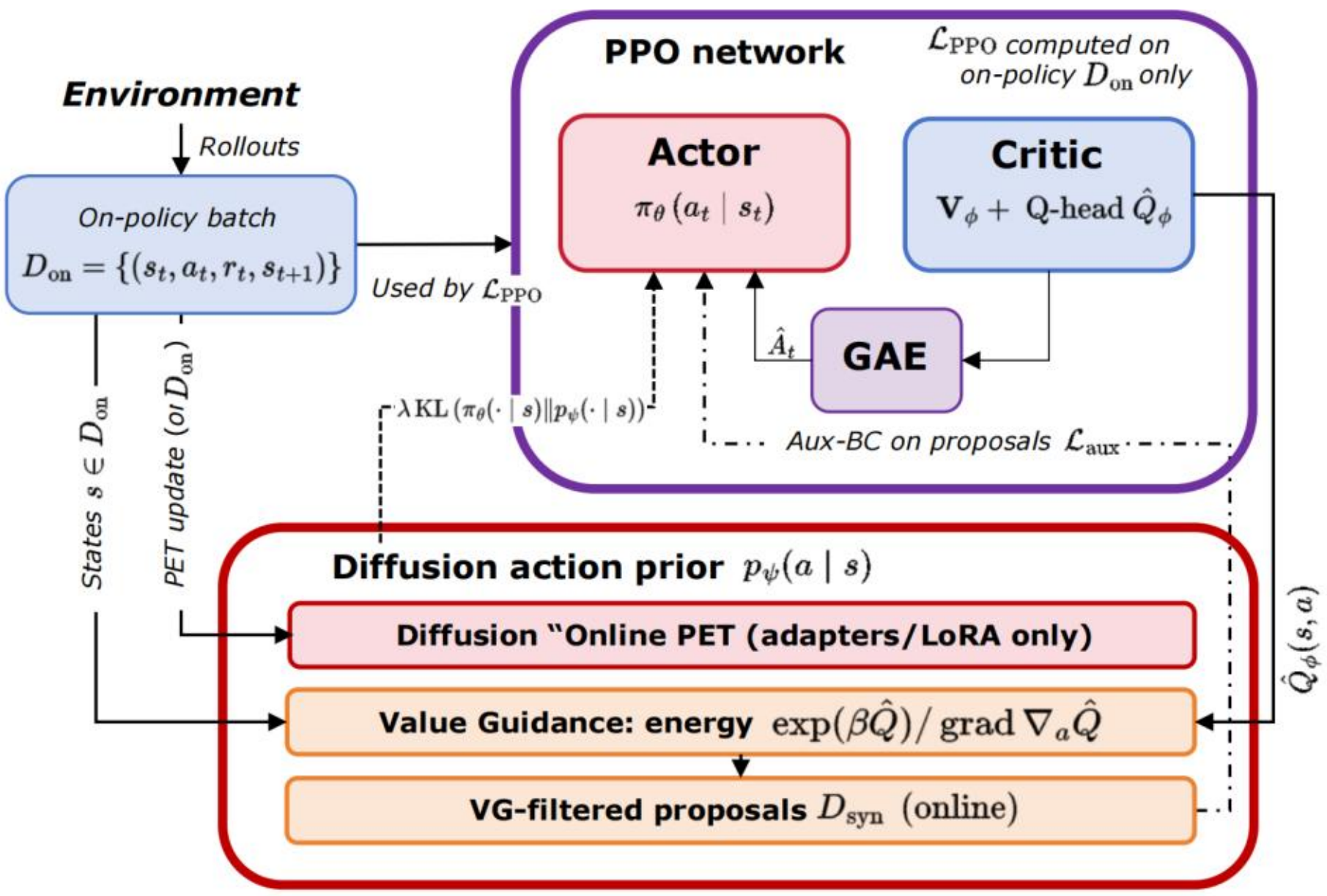

**Fig. 4.** Overall PPO-DAP training loop with a strictly on-policy PPO update and a value-guided diffusion prior. *At each iteration, on-policy data $D_{on}$ drive the PPO and GAE updates, while the diffusion action prior $p_\psi(a \mid s)$ generates value-guided proposals $D_{syn}$ ; for auxiliary losses and is updated via online PET on the same states; the "No-VG" and "No-PET" ablations respectively disable value guidance and PET adaptation.*

At online iteration $k$, we collect an on-policy batch $D_{\text{on}}^{(k)}$, compute $\hat{A}_t$ with GAE, and form $\mathcal{L}_{\text{PPO}}\left(D_{\text{on}}^{(k)}\right)$. The interaction between PPO, the diffusion prior, and value-guided proposals is summarized in Fig. 4. The overall actor objective is:

$$
\begin{aligned}
\mathcal{L}_{\text{actor}} &= \mathcal{L}_{\text{PPO}}\left(D_{\text{on}}^{(k)}\right) \\
&+\lambda_{\text{KL}} \mathbb{E}_{s\sim D_{\text{on}}^{(k)}}\left[\text{KL}\left(\pi_\theta(\cdot \mid s) \| p_\psi^G(\cdot \mid s)\right)\right] \\
&+\lambda_{\text{aux}}\, \mathbb{E}_{(s,a)\sim D_{\text{syn}}^{(k)}}\left[-\log\, \pi_\theta(a \mid s)\right].
\end{aligned}
\tag{9}
$$

The second term softly anchors the policy toward the prior (relative-entropy regularization is a standard stabilizer in policy search; Peters et al., 2010; Abdolmaleki et al., 2018). The third term is a lightweight imitation objective on value-guided proposals, consistent with maximum-likelihood / regression-style policy updates (Peng et al., 2019; Nair et al., 2020).

**Gaussian proxy head for tractable KL.** Since diffusion models do not provide a convenient closed-form density for KL computation, we introduce an auxiliary diagonal-Gaussian proxy head:

$$
p_\psi^G(\cdot \mid s) = \mathcal{N}\left(\mu_\psi(s), \text{diag}\left(\sigma_\psi^2(s)\right)\right),
\tag{10}
$$

this proxy is used **only** to compute and monitor the prior KL term, we compute $\mu_\psi(s)$ and $\sigma_\psi(s)$ by moment matching on the $K$ proposal actions sampled from the diffusion prior at state $s$. The resulting Gaussian $p_\psi^G$ is used solely for a tractable KL diagnostic/regularizer; it does not replace diffusion sampling, and (critically) we block gradients to $\psi$ through $\mathcal{L}_{\text{actor}}$ . The diffusion parameters are updated **only** through $\mathcal{L}_{\text{PET}}$.

**Key design points.**

1. PPO sampling, ratios, and advantages depend only on $D_{\text{on}}^{(k)}$;
2. $D_{\text{syn}}^{(k)}$ affects the actor only via small-weight auxiliary branches;
3. The critic shares a backbone: $V_\phi$ is trained with TD/GAE on $D_{\text{on}}^{(k)}$, and a lightweight $\hat{Q}_\phi(s, a)$ head (trained with one-step TD) is used only for proposal ranking/guidance.

## 4.5 Strict on-policy definition and online auditing

To make our "strictly on-policy" claim verifiable, we explicitly formalize the update boundary and implement runtime checks that audit data provenance and gradient flow.

**Definition 1 (Strictly on-policy update).** At iteration $k$, the policy update $\theta^{(k+1)} = \theta^{(k)} + \Delta\theta^{(k)}$ is *strictly on-policy* if:

(i) the PPO surrogate gradient $\nabla_\theta \mathcal{L}_{\text{PPO}}\left(\theta^{(k)}; \mathcal{D}_{\text{on}}^{(k)}\right)$ is computed solely from the freshly collected on-policy batch $\mathcal{D}_{\text{on}}^{(k)}$;

(ii) all auxiliary terms (the soft prior-KL regularizer and the auxiliary imitation loss) are evaluated only on states $s \in \mathcal{D}_{\text{on}}^{(k)}$ (i.e., every proposal tuple $(s,\cdot) \in D_{\text{syn}}^{(k)}$  $s \in D_{\text{on}}^{(k)}$);

(iii) the PET gradient $\nabla_{\psi_{\text{PET}}} \mathcal{L}_{\text{PET}}$ is computed only from $D_{\text{on}}^{(k)}$; and

(iv) the logged dataset $D_{\text{off}}$ is not used in any forward/backward pass that contributes to $\Delta\theta^{(k)}$ (i.e., no offline samples enter the computational graph of the actual policy update).

**Remark 1 (No offline gradient leakage).** Under Definition 1, the policy update $\Delta\theta^{(k)}$ is fully determined by $\left(\mathcal{D}_{\text{on}}^{(k)}, \psi^{(k)}\right)$ and does not depend $\mathcal{D}_{\text{off}}$; equivalently, offline logs are excluded from the policy-gradient pathway by construction.

**Implementation safeguards.** We enforce Definition 1 via simple engineering barriers:

- State provenance checks: we attach a state_id to each state in $D_{\text{on}}^{(k)}$ and propagate it to all proposal samples; we assert that all state_ids used by $\mathcal{L}_{\text{prior}}/\mathcal{L}_{\text{aux}}$ belong to the current batch.
- Stop-gradient isolation: prior outputs used in $\mathcal{L}_{\text{prior}}$ are explicitly detached to prevent PPO gradients from flowing into $\psi$.
- Dedicated PET dataloader: PET updates draw minibatches exclusively from $D_{\text{on}}^{(k)}$.
- Optimizer isolation: PPO and PET use separate optimizers (including momentum buffers) to avoid cross-stage interference.

**Online audit metrics.** We log three practical diagnostics:

- **Offline Gradient Leakage Ratio (OGLR).** Let $g_{\text{on}}^{(k)} = \nabla_\theta \mathcal{L}_{\text{actor}}$ denote the actor gradient used for the actual on-policy update at iteration $k$. In a *diagnostic-only* pass (no optimizer step), we run the same actor-gradient computation on an offline minibatch to obtain $g_{\text{off}}^{(k)}$. We define

$$\text{OGLR}^{(k)} = \frac{\left\| g_{\text{off}}^{(k)} \right\|_2}{\left\| g_{\text{on}}^{(k)} \right\|_2 + \delta}, \text{ ideal } \approx 0, \tag{11}$$

  where $\delta > 0$ is a small constant added to the denominator to avoid division by zero and improve numerical stability (e.g., when $\left\| g_{\text{on}}^{(k)} \right\|_2$ becomes extremely small). Ideally, $\text{OGLR}^{(k)} \approx 0$.

- **State Provenance Ratio (SPR)**. We measure the fraction of synthetic proposal samples whose states come from the current on-policy batch:

$$\text{SPR}^{(k)} = \frac{\#\left\{(s,\cdot) \in D_{\text{syn}}^{(k)} : s \in D_{\text{on}}^{(k)}\right\}}{\# D_{\text{syn}}^{(k)}}, \text{ ideal } = 1.0, \tag{12}$$

  Ideally, $\text{SPR}^{(k)} = 1.0$.

- **Policy-Gradient Share (PGShare)**. Let $g_{\text{PPO}}^{(k)} = \nabla_\theta \mathcal{L}_{\text{PPO}}\left(D_{\text{on}}^{(k)}\right)$ and $g_{\text{actor}}^{(k)} = \nabla_\theta \mathcal{L}_{\text{actor}}$. We define

$$\text{PGShare}^{(k)} = \frac{\left\| g_{\text{PPO}}^{(k)} \right\|_2}{\left\| g_{\text{actor}}^{(k)} \right\|_2 + \delta}, \text{ ideal } \approx 1. \tag{13}$$

  This metric is close to 1 when auxiliary regularization remains lightweight, as intended.

  Finally, we log policy-KL and prior-KL trajectories at each iteration to characterize the dual-proximal behavior in Section 5.

### 4.6 Training loop

**Training procedure.** Fig. 4 provides an overview of the online iteration with strictly on-policy PPO updates, value-guided proposal generation, and PET-based prior adaptation. For completeness and reproducibility, **Algorithm 1** summarizes the exact per-iteration steps and clarifies which losses update which parameter blocks under explicit gradient isolation.

---

Algorithm 1: PPO-DAP (Strictly on-policy with Value Guided Diffusion Prior)

---

Inputs: actor $\pi_\theta$, critic $V_\phi\left(+\hat{Q}_\phi\right)$ diffusion prior $p_\psi$ with PET params $\psi_{\text{PET}}$,
$\beta(\cdot)$ anneal, $\{\alpha_t\}$ guidance schedule, proposal count K

Repeat for iterations $k = 1..K_{max}$:

1) Rollout with $\pi_\theta$ to collect on-policy batch $\mathcal{D}_{\text{on}}^{(k)}$.
2) Compute $\hat{A}_t$ with GAE; form $\mathcal{L}_{\text{PPO}}\left(\mathcal{D}_{\text{on}}^{(k)}\right)$.
3) For each $s \in \mathcal{D}_{\text{on}}^{(k)}$:
   - Sample $\{a_1..a_K\} \sim p_\psi(\cdot|\ s)$// proposals (detached)
   - (optional) In-process guidance using Eq. (8) with $\{\alpha_t\}$
   - Energy weighting $w_i \propto \exp\ \left(\beta \hat{Q}_\phi(s, a_i)\right)$, resample $\Rightarrow \mathcal{D}_{\text{syn}}^{(k)}$ (Eq. (7)
4) Actor update (Eq. (9):

   $\theta \leftarrow \theta - \eta_\theta \nabla_\theta \mathcal{L}_{\text{actor}}$
   (gradients w.r.t. $\psi$ are blocked in $\mathcal{L}_{\text{actor}}$ )
   Critic update: $V_\phi$ via TD/GAE on $\mathcal{D}_{\text{on}}^{(k)}$; $\hat{Q}_\phi$ via one-step TD (shared trunk).
5) PET update (adapters/LoRA only):
   $\psi_{\text{PET}} \leftarrow \psi_{\text{PET}} - \eta_{\text{PET}} \nabla_{\psi_{\text{PET}}} \mathcal{L}_{\text{PET}}$*(denoising objective on* $\mathcal{D}_{\text{on}}^{(k)}$*)*
6) Monitoring: policy-KL, prior-KL, OGLR/SPR/PGShare

---

### 4.7 Complexity and cost

To avoid symbol overload, we denote the number of denoising steps in the diffusion sampler by $N_{\text{steps}}$ (the evaluation horizon $T$ used by ALC appears later in Section 5). Let $K$ be the number of proposals per state, $f$ the PET update frequency (e.g., PET steps per 100 actor updates), and $r$ the LoRA/Adapter rank.

- **Sampling cost:** diffusion sampling requires $\mathcal{O}\left(KN_{\text{steps}}\right)$ forward passes. Energy reweighting adds $\mathcal{O}(K)$. In-process gradient guidance adds extra $\mathcal{O}\left(KN_{\text{steps}}\right)$ evaluations of $\nabla_a \hat{Q}_\phi$.
- **Prior KL:** the Gaussian proxy head is $\mathcal{O}(1)$ per state and does not backpropagate to $\psi$ through $\mathcal{L}_{\text{actor}}$ .
- **PET updates:** backpropagation is restricted to adapters/LoRA, with approximate overhead $\mathcal{O}(f \cdot r)$ (Houlsby et al., 2019; Hu et al., 2021; Xie et al., 2023).

- **Memory:** scales roughly linearly with $K$ and whether gradient guidance is enabled. PET adds a small memory overhead due to low-rank trainable parameters.

If further speedups are required, one may replace the sampler with higher-order fast solvers to reduce $N_{\text{steps}}$ to 10–20, significantly lowering wall-clock proposal generation cost (Lu et al., 2022a; Lu et al., 2022b).

### 4.8 A dual-proximal performance lower bound

Conceptually, each PPO-DAP update can be viewed as a standard PPO on-policy update augmented with two small regularizers (prior KL and auxiliary imitation), while PET controls the drift of the prior distribution. We provide an informal lower bound in the spirit of monotonic improvement analyses for TRPO/PPO (Kakade & Langford, 2002; Schulman et al., 2015; Schulman et al., 2017).

**Proposition 1** (Informal, dual-proximal lower bound)

Assume: (i) the advantage function $A^{\pi_k}(s,a)$ is bounded; (ii) both the policy KL $\mathrm{KL}(\pi_{k+1} \| \pi_k)$ and the prior KL $\mathrm{KL}(p_{k+1} \| p_k)$ under the on-policy state distribution are bounded; and (iii) the value estimator $\hat{Q}_\phi$ used for guidance has bounded error on visited state–action pairs, e.g., $| \hat{Q}_\phi - Q^{\pi_k} | \leq \eta$. Then there exist constants $c_1, c_2, c_3 > 0$ such that

$$
\begin{aligned}
J(\pi_{k+1}) - J(\pi_k) & \\
&\geq L_{\pi_k}(\pi_{k+1}) - c_1 \mathrm{KL}(\pi_{k+1} \| \pi_k) - c_2 \mathrm{KL}(p_{k+1} \| p_k) \\
&\quad - c_3 \eta, \eta.
\end{aligned}
\tag{14}
$$

Here $L_{\pi_k}(\pi_{k+1})$ is the standard TRPO/PPO-style on-policy surrogate; $\mathrm{KL}(\pi_{k+1} \| \pi_k)$is primarily controlled by PPO's clipping/KL target; $\mathrm{KL}(p_{k+1} \| p_k)$ is controlled by small-step PET updates in a low-rank adapter subspace; and $\eta$ captures value-guidance error.

**Interpretation.** Eq. (14) suggests that expected improvement is dominated by the PPO surrogate term as long as: (a) policy updates remain proximal, (b) prior drift is kept small by PET, and (c) the value-guidance signal is not excessively biased. In practice, we do not optimize Eq. (14) directly; instead, we treat it as a sanity-check lens and monitor $\Delta J$ together with policy-KL and prior-KL trajectories. A proof sketch and the required technical conditions are provided in Appendix A.

## 5. Experiments

This section evaluates PPO-DAP under a strictly aligned online interaction budget and matched compute settings. We aim to answer three questions:

- **Effectiveness:** Can a diffusion action prior learned from logged data improve early exploration and final performance while keeping PPO strictly on-policy?
- **Ablations:** Under compute-fair comparisons, which components—value-guided proposals (VG) and parameter-efficient tuning (PET)—are primarily responsible for the gains?
- **Robustness:** How sensitive is PPO-DAP to logged-data coverage and critic quality, and what practical failure modes and mitigation strategies arise?

## 5.1 Setup and evaluation protocol

### 5.1.1. Tasks and environments

We evaluate on eight MuJoCo continuous-control tasks: Ant-v2, HalfCheetah-v2, Hopper-v2, HumanoidStandup-v2, Pusher-v2, Striker-v2, Swimmer-v3, and Walker2d-v3 (denoted as "Walker2d" when unambiguous). Each task uses an identical online interaction budget of $1.0 \times 10^6$ environment steps (≈200 epochs, with 1 epoch = 5k steps). Fig. 5 summarizes the task suite and evaluation schematic.

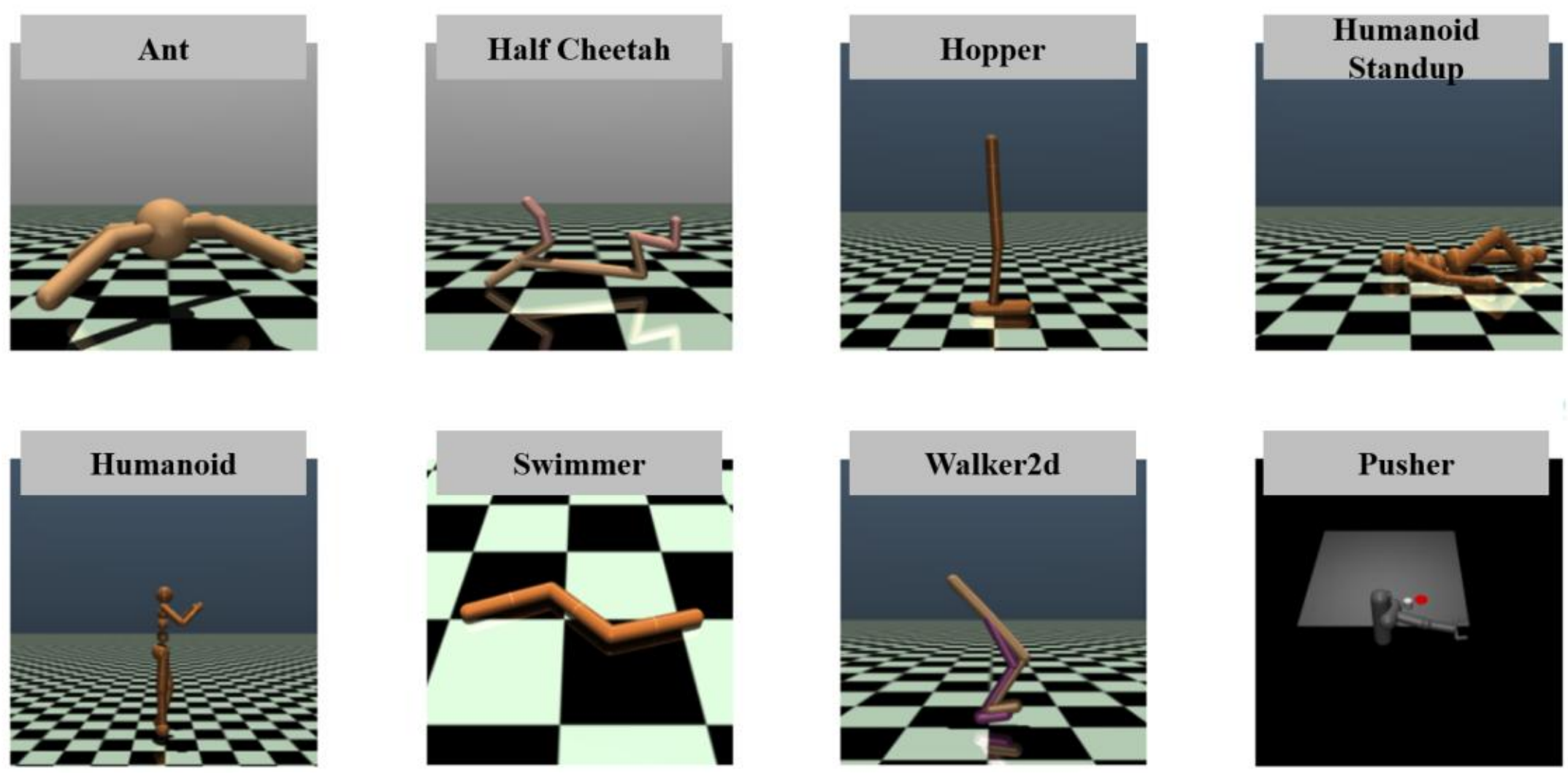

**Fig. 5. Task suite and evaluation protocol.** Each task uses $1.0 \times 10^6$ online environment steps with periodic evaluations; early learning efficiency is summarized by ALC@40 (first 40 epochs ≈ 200k steps).

### 5.1.2. Logged-data sources and the two-stage protocol

We follow the two-stage protocol in Sections 3–4. The logged dataset $D_{\text{off}}$ is used **only** to train the diffusion action prior $p_{\psi}(a \mid s)$ (and optionally to warm-start the actor–critic via supervised losses). No advantages are computed on logged trajectories and no PPO updates are performed on $D_{\text{off}}$. During online learning, PPO updates are computed **only** from freshly collected on-policy rollouts $D_{\text{on}}$, and the diffusion prior is adapted using PET on the same on-policy batch (Sections 4.2–4.5).

**Clarification on D4RL-style names.** When available, $D_{\text{off}}$ is taken from D4RL and we adopt the D4RL naming convention (e.g., *walker2d-medium-expert-v2*) solely to specify the source of logged data; all online evaluations are still conducted in the corresponding MuJoCo environment using newly collected on-policy rollouts.

### 5.1.3. Metrics and statistical testing

All strictly on-policy methods share the same online environment-step budget, network architectures, optimizer settings, and evaluation frequency. We evaluate every 5k–10k environment steps using a deterministic policy (mean action) and average returns over multiple trajectories. Each result uses 3–5 random seeds and is reported as mean ± 95% confidence interval (Student's t).

To summarize early learning efficiency, we report the Area under the Learning Curve over a horizon $T$:

$$\text{ALC@}T = \frac{1}{T}\int_0^T R(t)\mathrm{d}t$$

approximated using the trapezoidal rule over discrete evaluation points. Unless stated otherwise, we use $T = 40$ epochs (≈200k environment steps), denoted as ALC@40. For selected comparisons against PPO, we additionally report paired Wilcoxon signed-rank tests (matched seeds; $p < 0.05$).

### 5.1.4. Implementation details and hyperparameters

Experiments are run on an NVIDIA RTX 3090 (24 GB) with an Intel Xeon CPU. PPO uses a fixed recipe (clip $\epsilon = 0.2$, GAE $\lambda = 0.95$, batch size 256, and a $1.0 \times 10^6$ step budget). Unless stated otherwise, PPO-DAP uses $K = 10$ proposals per state and $N_{\text{steps}} = 20$ denoising steps, with energy reweighting and optional in-process gradient guidance for VG. PET updates only Adapter/LoRA parameters with learning rate $1 \times 10^{-5}$ and frequency $f \in \{0,5,10,20\}$. The prior-KL and auxiliary imitation losses are assigned small weights ($\lambda_{\text{KL}} \approx 5 \times 10^{-3}, \lambda_{\text{aux}} \approx 10^{-2}$) to keep PPO as the dominant update signal. To avoid duplicating implementation details across subsections, we summarize the default hyperparameters (and ablation ranges) in Table 2.

**Table 2.** Method-specific hyperparameters and online interaction budget.

| Component | Hyperparameter | Default (range) | Notes |
|---|---|---|---|
| Online budget | Environment steps / task | $1.0 \times 10^6$ | Shared across methods (≈200 epochs; 1 epoch = 5k env steps). |
| PPO | Clip parameter $\epsilon$ | 0.2 | See Eq. (5). |
| | GAE parameter $\lambda$ | 0.95 | Used for advantage estimation. |
| | Batch size | 256 | — |
| | Policy-KL monitoring target | 0.02 | If exceeded, reduce learning rate / epochs (monitoring/safety rule only). |
| Diffusion prior | Proposals per state (K) | 10 (5–20) | Used only by prior-KL / auxiliary branches; **no PPO gradients** are computed on proposals. |
| | Sampler steps $N_{\text{steps}}$ | 20 (10–30) | Number of denoising steps. |
| | Proposal generation frequency | once / iteration | Ablated in Fig. 9 (proposal count/frequency study). |
| Value guidance (VG) | Energy temperature $\beta$ | $0 \rightarrow 1.0$ | Linear annealing during the first 30% of training. |
| | Gradient guidance $\alpha_{max}$ | 0.30 (0–0.30) | See Eq. (8); tied to the noise schedule. |
| | $\left\Vert\nabla_a \hat{Q}_\phi\right\Vert_2$ cap $c$ | 0.1 | Prevents collapse / overly aggressive guidance. |
| Soft anchoring | $\lambda_{\text{KL}}$ | $5 \times 10^{-3}$ $(10^{-3} - 10^{-2})$ | Weight of the soft KL-to-prior term in Eq. (9); applied to the **actor only**. |
| Auxiliary imitation | $\lambda_{\text{aux}}$ | $1 \times 10^{-2}$ $(0 - 2 \times 10^{-2})$ | Auxiliary BC on $D_{\text{syn}}$ (value-guided proposals). |
| PET | PET update frequency $f$ | 10 (0/5/10/20) | PET steps per 100 actor updates; ($f = 0$) corresponds to **No-PET**. |
| | LoRA/Adapter rank $\tau$ | 8 (4–16) | Low-rank, proximal, compute-efficient adaptation. |

| | PET learning rate | $1 \times 10^{-5}$ | Update Adapters/LoRA only; diffusion backbone frozen online. |
|---|---|---|---|
| Protocol | Proposal ratio | $\leq 20\%$ | Caps proposal samples in an actor batch so auxiliary terms do not dominate PPO. |

## 5.2 Baselines and controls

To address concerns about “adding components or compute,” we compare against a set of baselines under the same online interaction budget ( $1.0 \times 10^6$ environment steps) and aligned compute settings. **Table 3 summarizes all baselines in a capability matrix**, including the strict on-policy boundary, whether logged data are used for advantage/ratio computation, the type of prior, and the use of value guidance (VG) and PET. We group the baselines into four categories: (i) strictly on-policy baselines and ablations; (ii) prior/control baselines; (iii) on-policy PPO variants; and (iv) offline and offline-to-online references (reported for context only).

**Table 3.** Baselines and controls (capability matrix).

| Method | Strictly on-policy?* | Advantages on logged data? | Prior type | VG | PET | Extra losses (online) |
|---|---|---|---|---|---|---|
| PPO (Vanilla) | ✓ | ✗ | None | ✗ | ✗ | None |
| PPO + BC Warmup | ✓ | ✗ | Warm-start (not a prior) | ✗ | ✗ | Offline BC only; none online |
| PPO-DAP (Ours) | ✓ | ✗ | Diffusion action prior | ✓ (Energy + Grad) | ✓ | Soft prior-KL + auxiliary BC (low weight) |
| PPO-DAP (No-VG) | ✓ | ✗ | Diffusion action prior | ✗ | ✓ | Soft prior-KL + auxiliary BC |
| PPO-DAP (No-PET) | ✓ | ✗ | Diffusion prior (frozen) | ✓ (Energy + Grad) | ✗ | Soft prior-KL + auxiliary BC |
| PPO-DAP (Prior-KL-only) | ✓ | ✗ | Diffusion action prior | ✗ | ✓ | Soft prior-KL only |
| PPO-DAP (Aux-only) | ✓ | ✗ | Diffusion action prior | ✓ (Energy / optional Grad) | ✓ | Auxiliary BC only |
| PPO + Behavior-prior KL | ✓ | ✗ | BC policy $\pi_{\mathrm{BC}}$ | ✗ | ✗ | $\mathrm{KL}(\pi_\theta \| \pi_{\mathrm{BC}})$ |
| PPO + Flow-prior | ✓ | ✗ | Flow-based prior | Energy-only† | ✗ | Prior-KL (+ optional aux) |
| PPO + AWR/AWAC-aux | ✓ | ✗ | None | ✗ | ✗ | Advantage-weighted BC (temperature grid) |

**Notes.** ✓ indicates “Yes/Enabled” and ✗ indicates “No/Disabled”. Strictly on-policy means that PPO importance ratios and advantages are computed only from the current on-policy rollouts $D_{\mathrm{on}}$ ; the logged dataset $D_{\mathrm{off}}$ is not used for advantage estimation or actor-gradient computation (Definition 1; audit metrics in §4.5). For fairness and stability, flow/VAE-prior controls use energy reweighting only (Eq. (7)) and disable in-denoising gradient guidance (Eq. (8)), since guidance quality can otherwise confound comparisons.

## 5.3 Main results

### 5.3.1. Overall findings

Under the unified online budget and aligned compute configuration, PPO-DAP achieves the best (or tied-best) final return on 6 out of 8 MuJoCo tasks and improves early learning efficiency on most tasks as measured by ALC@40. In the two tasks where PPO-DAP is not the strongest, HalfCheetah-v2 is nearly tied with the best on-policy baseline, while Striker-v2 exhibits sparse rewards and narrow logged-data coverage (Section 5.6), limiting the benefits of value-guided proposals.

### 5.3.2. Cross-task final return

**Table 4** reports the final episodic returns after $1.0 \times 10^6$ online environment steps. PPO-DAP matches or outperforms all strictly on-policy baselines on **Ant-v2, Hopper-v2, HumanoidStandup-v2, Swimmer-v3, and Walker2d-v3**, and remains competitive on **Pusher-v2**.

**Table 4.** Final return (mean ± 95% CI) after $1.0 \times 10^6$ environment steps. PPO-DAP denotes our method.

| TASK | PPO | PPO-DAP | PPO-GC | PPO-ARC | ABPPO |
|---|---|---|---|---|---|
| Ant-v2 | 183.5 ± 102.0 | **211.4 ± 114.4** | 196.0 ± 94.9 | 211.1 ± 122.9 | 182.1 ± 108.7 |
| Halfcheetah-v2 | 882.6 ± 149.5 | 1027.1 ± 112.0 | 792.7 ± 109.1 | **1027.2 ± 124.5** | 871.8 ± 113.0 |
| Hopper-v2 | 1429.5 ± 101.0 | **1542.9 ± 113.4** | 1453.3 ± 90.4 | 1532.9 ± 121.7 | 1423.8 ± 109.9 |
| Humanoidstandup-v2 | 80524.7 ± 91.7 | **83174.0 ± 101.1** | 79885.4 ± 97.8 | 82174.7 ± 118.0 | 82096.3 ± 107.9 |
| Pusher-v2 | −52.7 ± 93.3 | **−51.1 ± 115.1** | −51.4 ± 101.2 | **−51.1 ± 117.5** | −54.2 ± 110.9 |
| Striker-v2 | −253.6 ± 103.0 | −227.3 ± 117.0 | −253.9 ± 106.4 | **−219.3 ± 100.9** | −249.0 ± 105.5 |
| Swimmer-v3 | 84.2 ± 91.0 | **95.4 ± 114.8** | 89.6 ± 103.7 | 94.0 ± 106.4 | 95.1 ± 113.2 |
| Walker2d-v3 | 766.5 ± 98.6 | **906.3 ± 102.3** | 664.1 ± 95.9 | 776.3 ± 100.4 | 905.3 ± 103.0 |

**Notes.** Values are reported as mean ± 95% confidence interval (CI) across seeds (Student's *t*). For negative-return tasks (Pusher/Striker), "higher is better" still holds (less negative).

### 5.3.3. Early efficiency and significance testing

**Table 5** focuses on **early learning efficiency (ALC@40 epochs)** and **final return** on four representative settings and reports paired **Wilcoxon signed-rank tests** against PPO (matched seeds). PPO-DAP improves both ALC@40 and final return across the four settings, and reaches statistical significance on the **Walker2d-medium-expert** setting ($p < 0.05$); the remaining settings show consistent but weaker trends.

**Clarification on task naming and logged-data source.** In Table 5 (and Figs. 6–11), tasks with **D4RL-style names** (e.g., "medium-expert", "medium-replay") indicate that $D_{\text{off}}$ comes from a D4RL-style logged dataset, while tasks named by standard MuJoCo IDs (e.g.,

**HumanoidStandup-v2**) use **self-collected logged data** for $D_{\text{off}}$. In all cases, the **online training and PPO updates remain strictly on-policy**.

**Table 5.** Early learning efficiency (ALC@40 epochs) and final return on representative tasks (mean ± 95% CI). Δ denotes the relative improvement over PPO, and $p$ is from a paired Wilcoxon signed-rank test (matched seeds).

**(a) ALC@40 (epochs)**

| Task | ALC@40 (PPO) | ALC@40 (PPO-DAP) | Δ | *p*-value |
|---|---|---|---|---|
| **Walker2d-medium-expert (D4RL)** | 1,900 ± 90 | 2,050 ± 85 | 7.90% | 0.038 |
| **HalfCheetah-medium-replay (D4RL)** | 1,800 ± 80 | 1,925 ± 85 | 6.90% | 0.071 |
| **Hopper-medium-expert (D4RL)** | 1,050 ± 70 | 1,100 ± 75 | 4.80% | 0.114 |
| **HumanoidStandup-v2 (online)** | 74,000 ± 1,800 | 75,500 ± 1,700 | 2.00% | 0.089 |

**(b) Final return**

| Task | Final return (PPO) | Final return (PPO-DAP) | Δ | *p*-value |
|---|---|---|---|---|
| **Walker2d-medium-expert (D4RL)** | 2,850 ± 110 | 3,030 ± 100 | 6.30% | 0.049 |
| **HalfCheetah-medium-replay (D4RL)** | 2,350 ± 120 | 2,470 ± 130 | 5.10% | 0.092 |
| **Hopper-medium-expert (D4RL)** | 1,500 ± 90 | 1,560 ± 100 | 4.00% | 0.122 |
| **HumanoidStandup-v2 (online)** | 80,600 ± 1,000 | 82,300 ± 1,100 | 2.10% | 0.065 |

### 5.3.4. Learning curves and cross-task visualization

**Figure 6** presents representative learning curves (e.g., **Walker2d-medium-expert-v2**, where $D_{\text{off}}$ is drawn from D4RL-style logged data). PPO-DAP exhibits a noticeably steeper improvement after ∼10–16 epochs and reaches a higher plateau around epochs 30–40, indicating gains in both early sample efficiency and asymptotic performance.

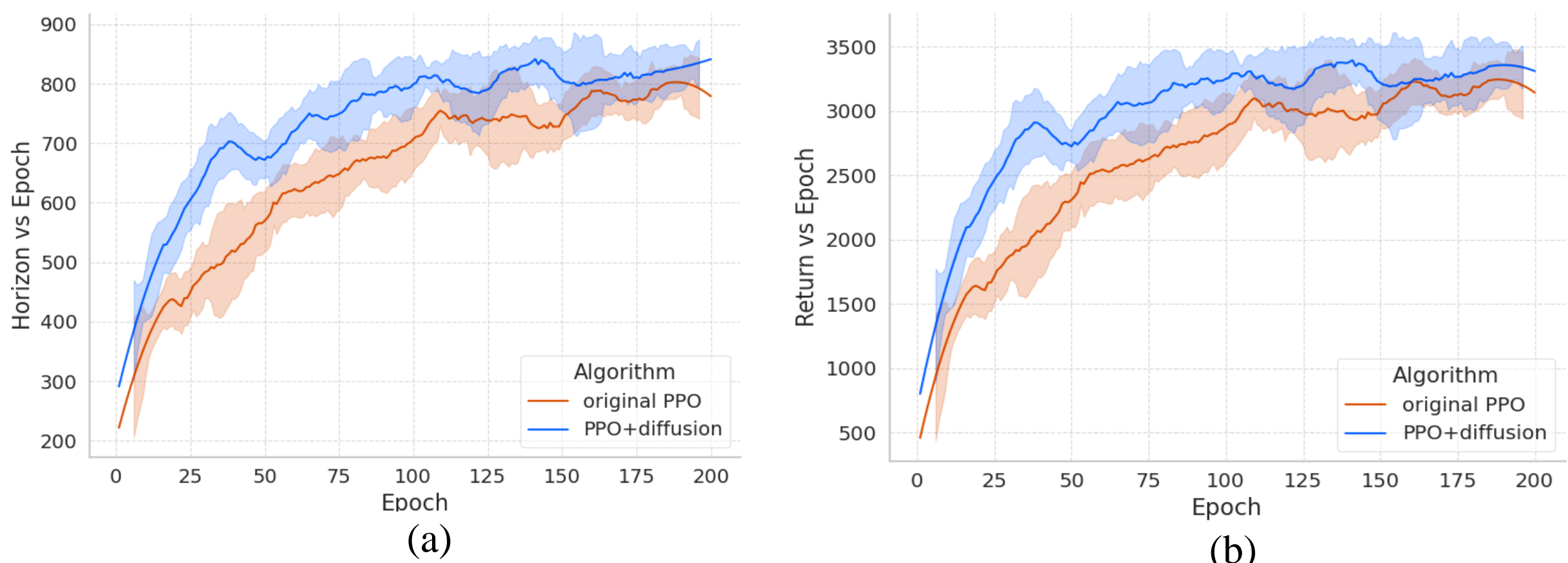

**Fig. 6. PPO vs. PPO-DAP on Walker2d-medium-expert-v2.** (a) Episode horizon (length) versus environment steps. (b) Episodic return versus environment steps. Curves show mean performance and shaded regions denote 95% confidence intervals across seeds.

**Figure 7** further summarizes cross-task improvements via a forest plot for ΔALC@40 and ΔFinal return (relative to PPO). For tasks with negative PPO returns (Pusher/Striker), ΔFinal is normalized by |PPO| for interpretability.

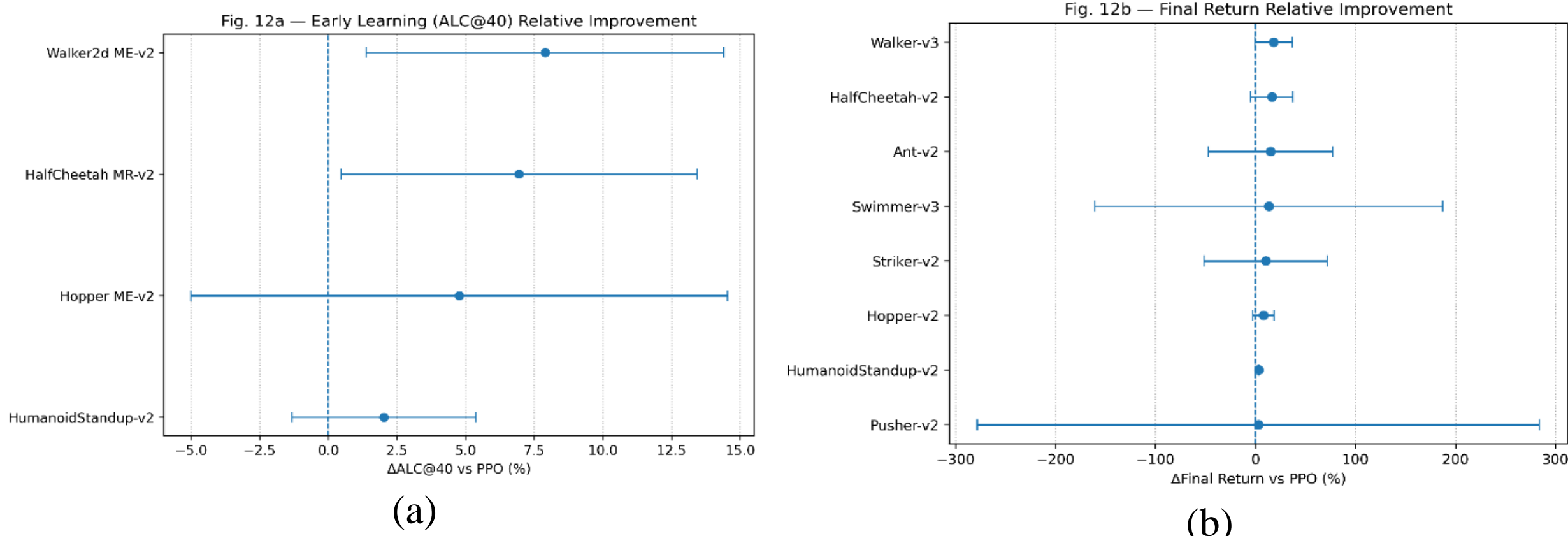

(a) (b)

**Fig. 7. Cross-task relative improvements of PPO-DAP over PPO.** (a) Relative gain in early learning efficiency, ΔALC@40 (%), on D4RL-based settings. (b) Relative gain in final return, ΔFinal (%), across tasks; for negative-return baselines (e.g., Pusher/Striker), improvements are normalized by ∣ PPO ∣ for interpretability. Error bars indicate 95% confidence intervals of paired differences.

**Anomalies and qualitative explanations.**

- **Striker-v2:** Sparse rewards and narrow logged-data coverage reduce the critic's ability to provide reliable value guidance early in training; consequently, PPO-DAP's relative gains are limited (see the coverage scan in Section 5.6).
- **HalfCheetah-v2:** PPO-DAP is nearly tied with PPO-ARC, suggesting that in smooth-reward, comparatively easier dynamics, PPO-DAP is at least competitive with strong on-policy objective modifications.

## 5.4 Compute fairness and cost

### 5.4.1. Measurement protocol

To assess whether the gains of PPO-DAP are attributable to algorithmic improvements rather than additional compute, we quantify its overhead under strictly aligned experimental conditions. Specifically, all methods share the same hardware setup, number of environment workers, network architectures, optimizer settings, batch sizes, evaluation frequency, and the unified online interaction budget of $1.0 \times 10^6$ environment steps per task.

We report three complementary compute metrics:

1. Throughput (↑) the average number of environment steps processed per second over the training run, normalized by the PPO baseline.
2. Wall-clock time × PPO (↓): the total elapsed training time required to complete $1.0 \times 10^6$environment steps, normalized to vanilla PPO (= 1.00).
3. Peak GPU memory × PPO (↓): the maximum GPU memory consumption during training, normalized to PPO (= 1.00).

Wall-clock time is measured using a synchronized timer around the rollout + update loop, and peak memory is sampled with nvidia-smi at 2-second intervals and aggregated by the maximum observed value. These metrics are reported in Table 7 as cross-task averages (relative to PPO).

### 5.4.2. Controllable cost knobs

The additional overhead of PPO-DAP arises from two sources: **proposal generation** (diffusion sampling and optional guidance) and **online prior adaptation** (PET/PEFT updates). We expose five controllable "knobs" that govern this overhead:

- $K$: number of proposals per on-policy state;
- $N_{\text{steps}}$ : number of denoising steps per proposal;
- Grad-guided: whether in-process gradient guidance is enabled during denoising;
- $f$: PET update frequency (e.g., PET steps per 100 PPO actor updates);
- $r$: adapter/LoRA rank.

Knobs ( $K$ , $N_{\text{steps}} = 20$ , Grad-guided) primarily control **sampling-side** cost, whereas $(f, r)$ control the **backpropagation-side** cost of PET. The default configuration used for the main results is $K = 10$, $N_{\text{steps}} = 20$, gradient guidance enabled, $f = 10$, and $r = 8$.

### 5.4.3. Cost–performance trade-off

**Single-task analysis (Walker2d).**

Table 6 isolates the incremental cost of PET by normalizing compute metrics to the **No-PET** configuration (diffusion prior enabled but adapters frozen). Under this convention, moderate PET $(f = 10, r = 8)$ increases wall-clock time by 1.17 × and peak memory by 1.05 × relative to No-PET, while improving final return and early efficiency (ALC@40).

A key presentation point (to avoid reviewer confusion): **Table 6 uses mixed normalization**—compute columns are × No-PET, but the performance gain Δ is reported **relative to vanilla PPO**. This is already reflected by the "Δ vs PPO" columns in Table 6.

**Table 6.** PET overhead versus performance on Walker2d (mean over 3–5 seeds).
***Compute metrics*** *(wall-clock time and peak GPU memory) are normalized to the* ***No-PET*** *configuration (diffusion prior enabled, adapters/LoRA frozen) to isolate the incremental cost of PET.* ***Performance gains*** *Δ are reported* **relative to vanilla PPO**. *Here* $f$ *denotes the PET update frequency (PET steps per 100 PPO actor updates) and* $r$ *is the adapter/LoRA rank.*

(a) Overhead and final return

| Setting | Wall-clock time (× No-PET) | Peak GPU memory (× No-PET) | Final return | Δ vs PPO |
|---|---|---|---|---|
| PPO | — | — | 2,850 ± 110 | 0 |
| No-PET ($f = 0$) | 1.00× | 1.00× | 3,000 ± 100 | 5.30% |
| PET ($r = 8, f = 5$) | 1.10× | 1.03× | 3,040 ± 98 | 6.70% |
| PET ($r = 8, f = 10$) | 1.17× | 1.05× | 3,070 ± 95 | 7.70% |
| PET ($r = 8, f = 20$) | 1.22× | 1.08× | 3,090 ± 92 | 8.40% |

(b) ALC@40 and relative gains

| Setting | Δ vs No-PET | ALC@40 (epochs) | Δ vs PPO | Δ vs No-PET |
|---|---|---|---|---|
| PPO | — | 1,900 ± 90 | 0 | — |
| No-PET ($f = 0$) | — | 2,100 ± 85 | 10.50% | — |
| PET ($r = 8, f = 5$) | 1.30% | 2,120 ± 82 | 11.60% | 1.00% |
| PET ($r = 8, f = 10$) | 2.30% | 2,150 ± 80 | 13.20% | 2.40% |
| PET ($r = 8, f = 20$) | 3.00% | 2,170 ± 78 | 14.20% | 3.30% |

Notes. $f$: PET update frequency (PET steps per 100 PPO actor updates). $r$: LoRA/Adapter rank. "× No-PET" normalizes overhead metrics by the No-PET configuration (diffusion prior enabled but adapters frozen).

**Cross-task summary.** Table 7 reports compute fairness aggregated across 8 tasks and normalized to PPO. PPO-DAP (Full) yields an average wall-clock overhead of $1.18 \pm 0.04$ and a modest peak-memory overhead of $1.05 \pm 0.02$, with throughput reduced to $0.86 \pm 0.04$ relative to PPO. Disabling value guidance (No-VG) or PET (No-PET) reduces overhead but also shifts the method toward a less favorable compute–return trade-off.

**Table 7**. Compute-fairness summary averaged across eight tasks (normalized to PPO). *Higher throughput is better; lower wall-clock time and peak GPU memory are better. Values are mean ± dispersion across tasks.*

| Method | K | $N_{\text{steps}}$ | Grad-Guided? | PET $f$ | $r$ | Throughput↑ | Wall-clock ×PPO↓ | Peak-mem ×PPO↓ |
|---|---|---|---|---|---|---|---|---|
| PPO (Vanilla) | — | — | — | — | — | 1.00 ± 0.00 | 1.00 ± 0.00 | 1.00 ± 0.00 |
| PPO + BC Warmup | — | — | — | — | — | 0.99 ± 0.01 | 1.01 ± 0.01 | 1.00 ± 0.00 |
| PPO-DAP (Full) | 10 | 20 | ✓ | 10 | 8 | 0.86 ± 0.04 | 1.18 ± 0.04 | 1.05 ± 0.02 |
| PPO-DAP (No-VG) | 10 | 20 | ✗ | 10 | 8 | 0.90 ± 0.03 | 1.12 ± 0.03 | 1.03 ± 0.01 |
| PPO-DAP (No-PET) | 10 | 20 | ✓ | 0 | — | 0.91 ± 0.03 | 1.10 ± 0.03 | 1.03 ± 0.02 |
| Diffusion-NoVG (ctrl) | 10 | 20 | ✗ | 10 | 8 | 0.89 ± 0.03 | 1.14 ± 0.03 | 1.04 ± 0.01 |
| VAE + PPO (ctrl) | 10 | — | — | — | — | 0.95 ± 0.02 | 1.08 ± 0.02 | 1.02 ± 0.01 |

**Return–compute visualization.** Figure 8 summarizes the return–compute trade-off via a scatter plot and Pareto envelope. PPO-DAP (Full) typically lies on or near the Pareto front, indicating that the performance gains are achieved with a controlled overhead (about +18% wall-clock time and about +5% peak memory on average).

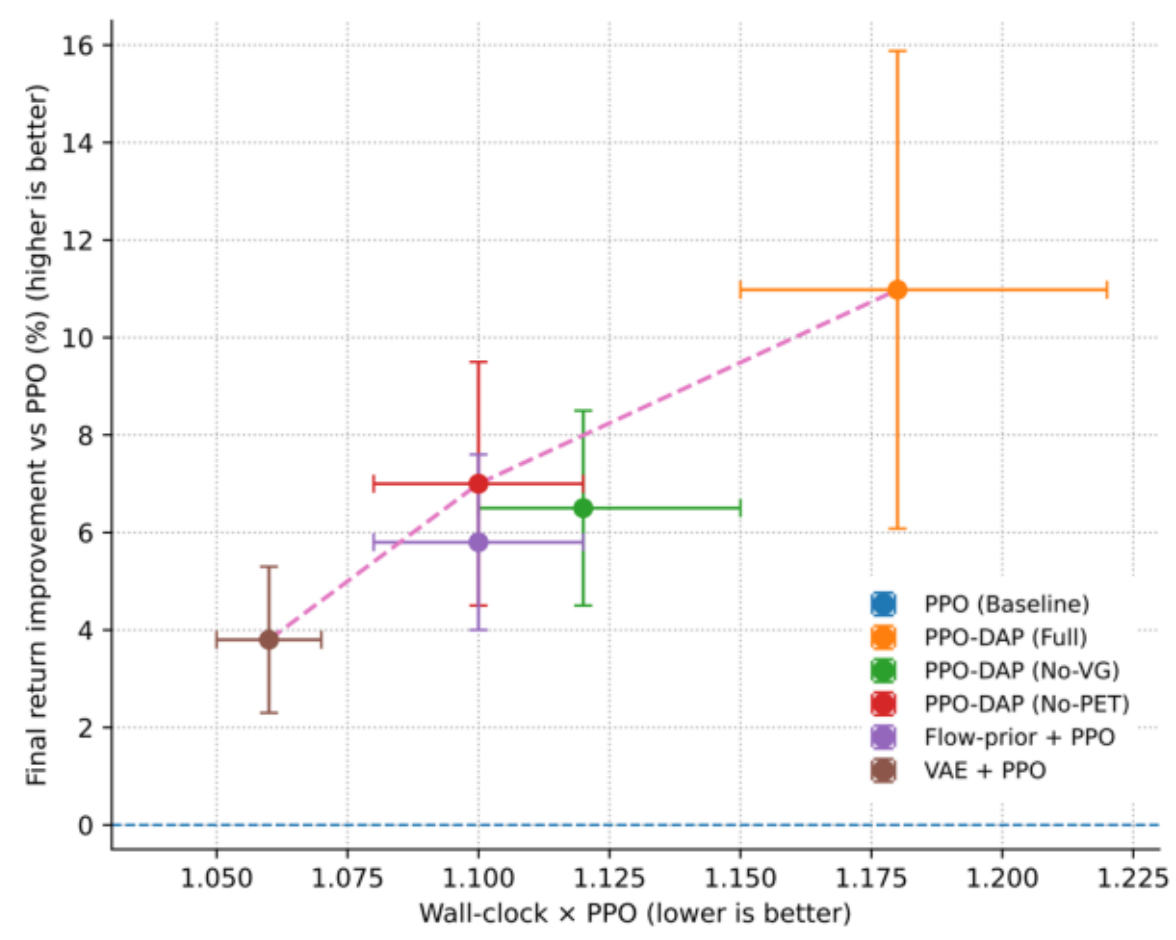

**Fig. 8.** Return–compute trade-off under compute-aligned settings.

### 5.5 Factorized ablations

**Goal.** We perform a factorized ablation to test whether the three key ingredients of PPO-DAP exhibit **non-interchangeable synergy**: (i) the **prior type** (none / flow / diffusion), (ii) the **value-guidance mechanism** (none / energy / energy + gradient), and (iii) the **online prior adaptation frequency** via PET$f \in \{0,10,20\}$. The goal is to verify that only a well-matched combination can simultaneously yield **faster early learning**, **higher final return**, and **controllable overhead**, rather than gains that could be replicated by swapping in a weaker component.

**Experimental protocol.** All ablations share the same online interaction budget ($1.0 \times 10^6$ environment steps), evaluation frequency, and default hyperparameters (Table 2), ensuring comparability. Unless stated otherwise, the ablations are conducted on **D4RL-style logged-data variants** (e.g., *Walker2d medium expert*, *HalfCheetah medium replay*, *Hopper medium expert*) while **keeping online updates strictly on-policy** during fine-tuning.

We vary the following factors:

- Prior type: no prior / flow-based prior / diffusion prior.
- Guidance:
    - Energy guidance (post-hoc reweighting with temperature $\beta$),
    - Energy + gradient guidance (in-denoising guidance with weight schedule $\alpha_t$).
- PET frequency: $f \in \{0,10,20\}$. where $f$ denotes the number of PET updates per 100 actor updates (and $f = 0$ is No-PET).

Controlled settings (unless ablated): proposal count $K = 10$, denoising steps $N_{\text{steps}} = 20$, $\beta$ linearly annealed from 0 to 1 over the first 30% of training, $\alpha_{max} = 0.30$, and action-gradient clipping $\left\|\nabla_a \hat{Q}_\phi\right\|_2 \leq 0.1$. For Flow/VAE/behavioral priors, we enable energy guidance only (no gradient guidance) to avoid confounding effects due to different gradient fidelity across prior classes; for the diffusion prior, we evaluate both Energy and Energy+Grad.

**Metrics.** We report early learning efficiency via ALC@40 (area under the learning curve over the first 40 epochs), final return, compute overhead (Wall-clock×PPO and Peak-mem×PPO), proposal "acceptance" (fraction of generated candidates retained after optional filtering/selection), and stability diagnostics (policy-KL and prior-KL trajectories).

**Key findings**

**(1) Proposal budget: $K$ and proposal-generation frequency**

As shown in **Fig. 9**, increasing either the number of proposals per state $K$ or the proposal-generation frequency accelerates early learning on tasks such as Hopper and Walker2d, and can also improve the final performance plateau on HalfCheetah. However, very large $K$ yields slightly higher variance in learning curves, suggesting a practical trade-off between proposal diversity/quality and training stability.

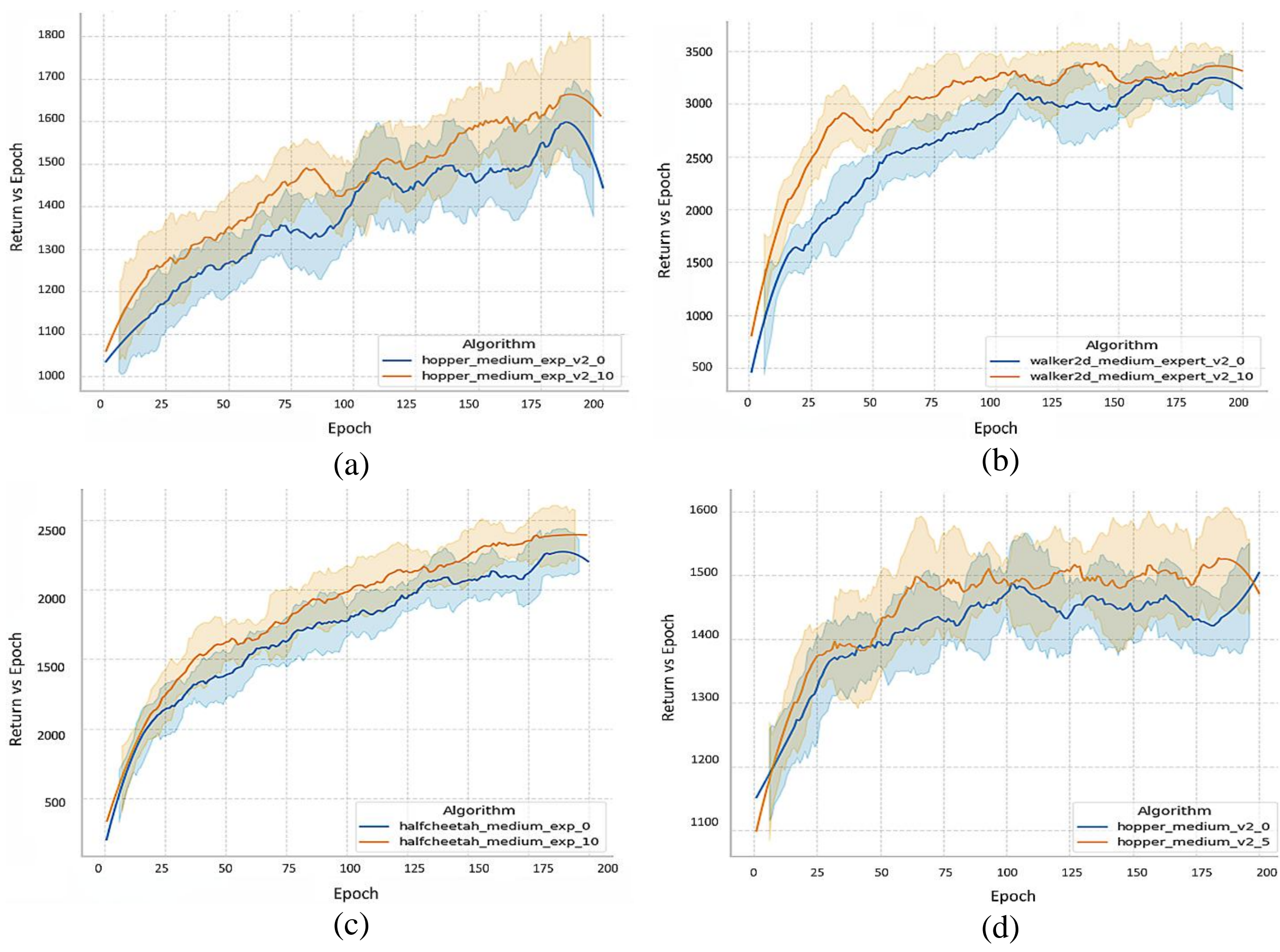

**Fig. 9.** *Ablation on proposal budget.* Increasing the number of proposals per state $K$ or the proposal-generation frequency accelerates early learning (e.g., Hopper/Walker2d) and can improve the final plateau (e.g., HalfCheetah). Extremely large $K$ slightly increases variance, indicating a practical operating region rather than "the more the better".

**(2) Online PET adaptation**

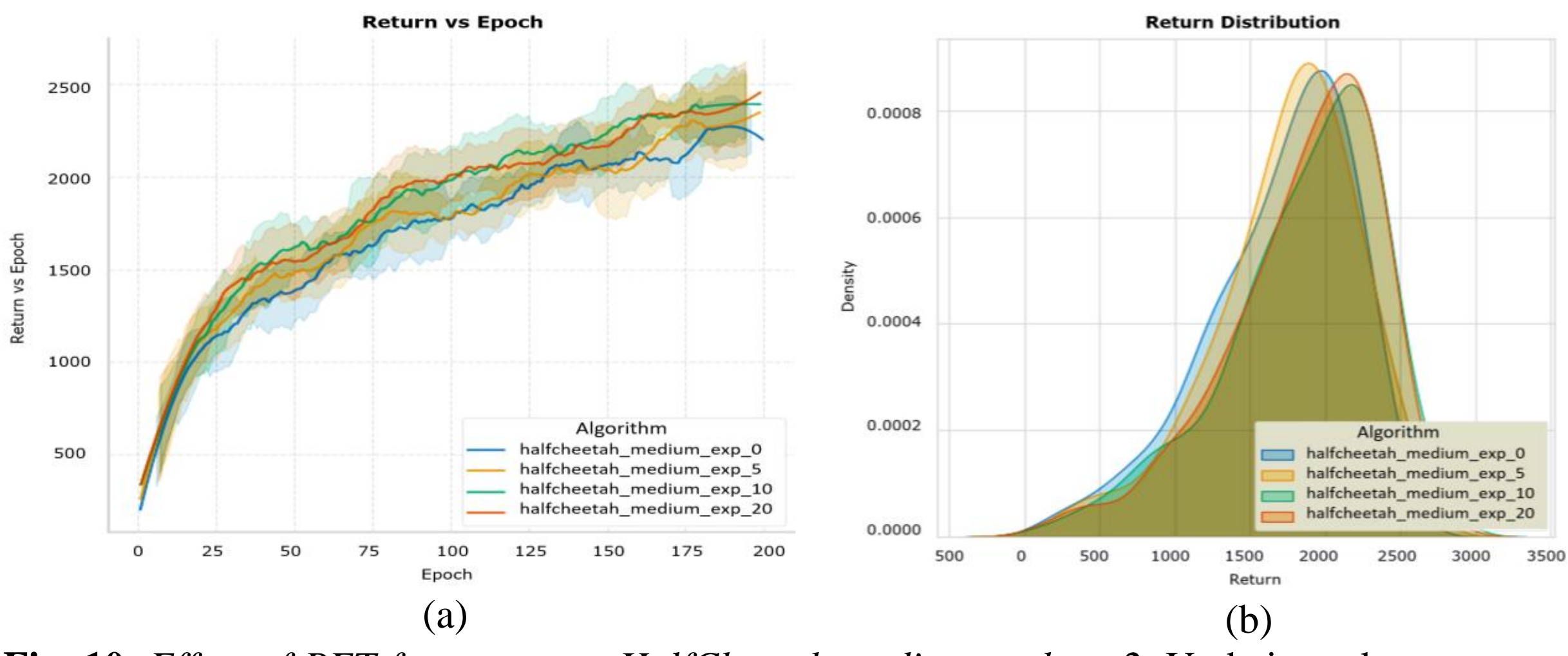

**Fig. 10.** *Effect of PET frequency on HalfCheetah medium replay v2.* Updating adapters every 10–20 actor-optimization intervals outperforms No-PET and low-frequency settings and shifts the final-return distribution to the right by ≈ 200–300. (a) Learning curves; (b) final return distribution.

**Fig. 10** studies the PET update frequency on *HalfCheetah medium replay v2*. Medium-to-high PET frequencies ($f \in [10,20]$) consistently outperform No-PET and low-frequency updates: after roughly 100k steps, performance separates clearly, and the distribution of final returns shifts right by approximately 200–300. Considering compute cost (Section 5.4), $f = 10$ provides the best performance–overhead trade-off, while $f = 20$ shows diminishing marginal gains with higher wall-clock cost.

**(3) Contribution of value guidance**

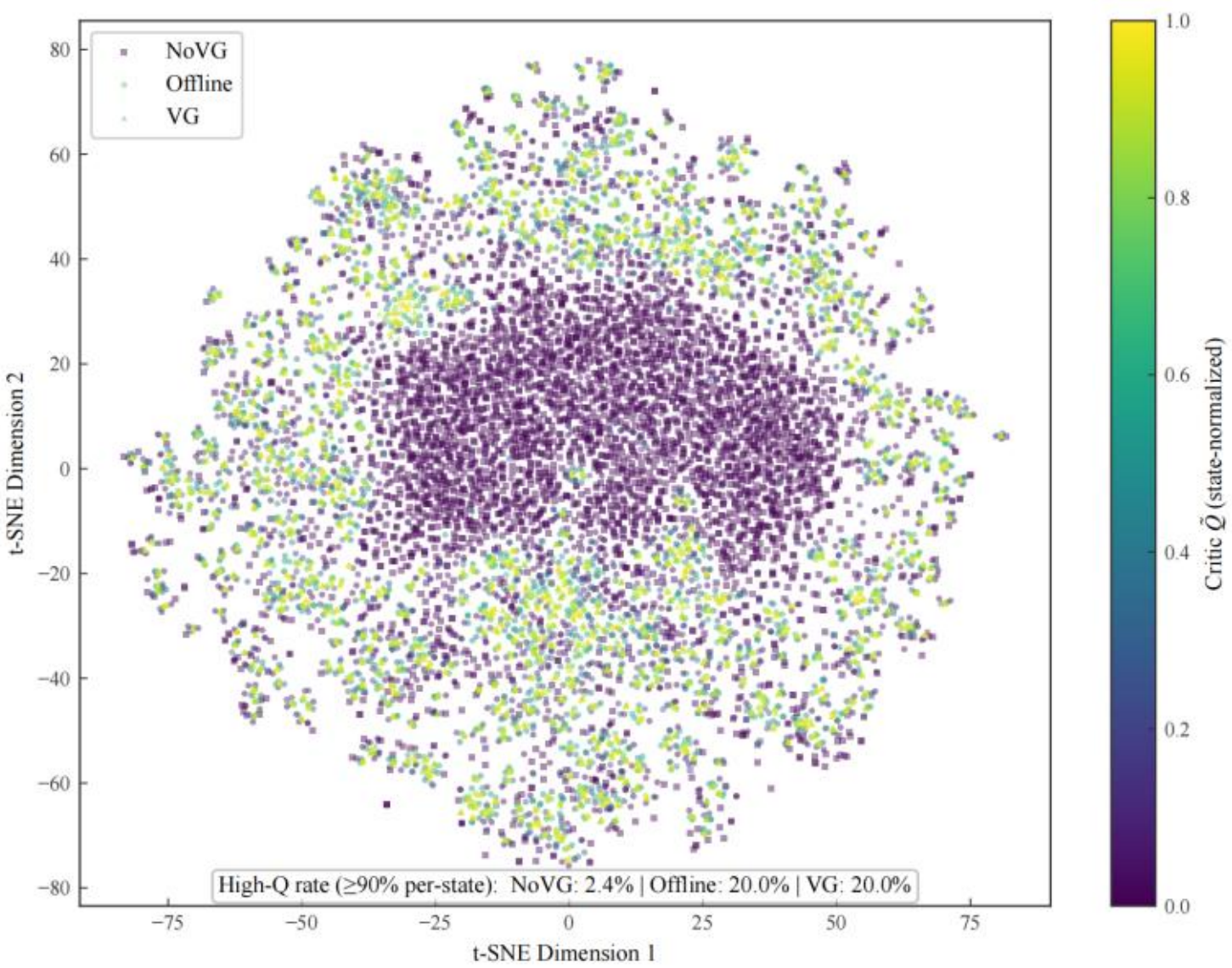

**Fig. 11.** *t-SNE of proposals on the same on-policy state batch.* Points are colored by the state-normalized critic score $\tilde{Q}$. Value guidance (Diffusion-VG) moves proposals toward higher-$\tilde{Q}$ regions and increases the within-state High-$\tilde{Q}$ fraction (≥ 90th percentile). The overlap among groups indicates that guided proposals remain consistent with the underlying action distribution rather than collapsing to a single mode.

**Fig. 11** visualizes proposals generated on the same on-policy state batch using t-SNE. Compared with unguided diffusion proposals (Diffusion-NoVG), **Energy+Grad** guidance concentrates proposals into regions with higher **state-normalized** critic scores $\tilde{Q}$, increasing the fraction of high - $\tilde{Q}$ candidates (≥ 90th percentile within each state) by roughly 30%–40. Energy-only guidance improves the proposal distribution but is less effective than Energy+Grad in both "focusing" and acceptance rate. Importantly, with $\beta$ -annealing and gradient-norm clipping, we do not observe obvious mode collapse.

**(4) Dual-proximal stability under aligned compute**

Across training, **prior-KL remains substantially smaller than policy-KL** (often by an order of magnitude), indicating that PET updates stay in a proximal regime. Together with the compute measurements in Table 6 (single-task) and Table 7 (cross-task), this supports that PPO-DAP's gains are achieved with **controllable overhead** rather than by unconstrained additional computation.

**Summary (C3: non-interchangeable synergy)**

Overall, the results suggest a clear interaction effect:

- A **diffusion prior without VG** mainly acts as a mild behavioral anchor and typically yields limited early-learning acceleration (consistent with "Prior-KL-only/Aux-only" controls).
- **Energy+Grad** further increases the high-value hit rate and effective learning signal beyond energy-only guidance, improving ALC@40 more reliably.
- **PET** is essential for tracking the drift between the offline-trained prior and the evolving on-policy state distribution; $f = 10$ is a robust default, while $f = 0$ leads to gradual mismatch and $f = 20$ yields diminishing returns at higher cost.

Consequently, **(Diffusion prior + Energy+Grad + moderate PET)** is the only configuration that consistently achieves "faster early learning + higher final return + controllable cost" in our study.

## 5.6 Coverage sensitivity

**Purpose.** We evaluate how PPO-DAP depends on the **support and diversity** of the logged dataset $D_{\text{off}}$, and we identify practical boundaries and mitigation strategies when logged coverage is limited.

**Protocol**

We construct three coverage levels during prior pretraining while keeping the online protocol fixed:

- **Wide:** full $D_{\text{off}}$.
- **Medium:** stratified subsampling that preserves the proportion of trajectories across return strata.
- **Narrow:** biased subsampling that removes high-return and tail segments, producing a more concentrated behavioral distribution with reduced action variance.

During online training, we keep $K, N_{\text{steps}}$, guidance settings, PET frequency, and evaluation protocol unchanged.

**Metrics.** We track:

- $\Delta$ALC@40: relative improvement in early learning efficiency.
- $\Delta$Final: relative improvement in final return.
- **High-$\boldsymbol{Q}$ hit rate:** $\Pr\left[\hat{Q}(s,a) \geq q_{0.9}(\hat{Q}(\cdot \mid s)\right)$, computed over proposals sampled at the same on-policy state $s$.
- **Effective sample size (ESS)** after energy reweighting: $\text{ESS} = 1/\sum_i w_i^2$, where $\{w_i\}$ are the normalized energy weights.

**Fig. 12(a–d)** reports these metrics under Wide/Medium/Narrow coverage.

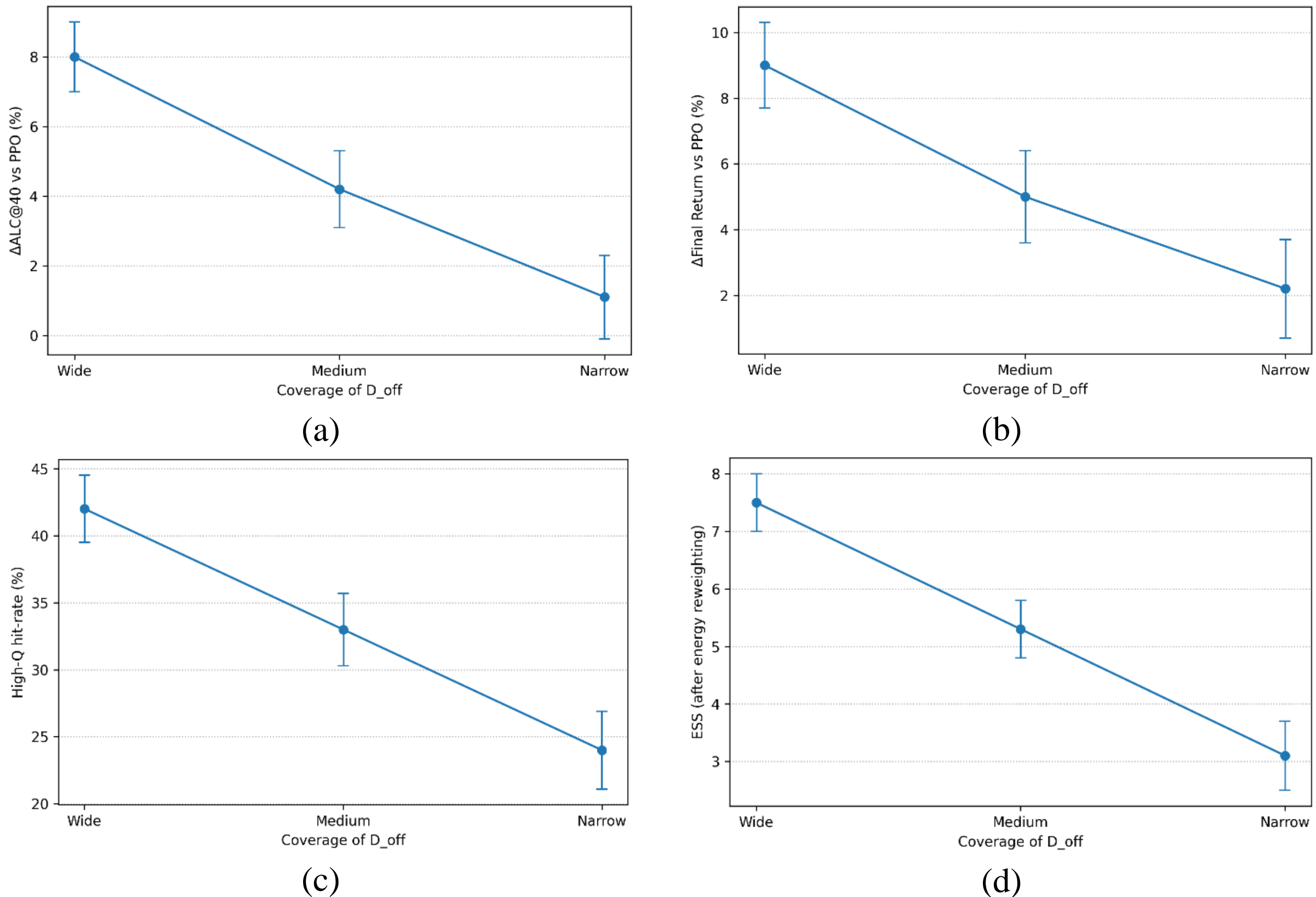

**Fig. 12.** *Coverage sensitivity.* As logged-data coverage narrows (Wide → Medium → Narrow), both ΔALC@40 and ΔFinal decrease, accompanied by reduced High - $Q$ hit rate and lower ESS after energy reweighting. Markers denote per-task/per-seed results; lines show mean ± 95% CI (bootstrap).

**Findings and practical mitigations.** We observe a monotonic degradation as coverage narrows: both early efficiency and final return improvements diminish, with larger drops in sparse-reward or high-dimensional settings. The concurrent decrease in High - $Q$ hit rate and ESS suggests that **proposal quality** and **weight degeneracy** are the primary bottlenecks.

For narrow-coverage logs, we recommend the following engineering mitigations:

- **Lower and gate** the energy temperature (e.g., $\beta_{max} \approx 0.6$) to avoid overly peaky reweighting early in training.
- **Increase $K$** moderately (e.g., $K = 15 - 20$) while using diversity-aware sampling to maintain proposal diversity.
- **Weaken gradient guidance** (e.g., $\alpha_{max} \approx 0.2$) while keeping gradient-norm clipping to prevent instability.
- Apply **lightweight $Q$-calibration** to reduce overconfidence and ranking artifacts.

Across conditions, $f = 10$ remains a robust default for performance–cost balance.

### 5.7 Reliability of value guidance

Value guidance depends on the accuracy of the critic-based action-value estimator $\hat{Q}_\phi(s,a)$. If $\hat{Q}_\phi$ is severely biased on visited state–action pairs, energy/gradient guidance may amplify errors and induce overly concentrated proposals.

We assess VG reliability using:

- **Spearman rank correlation $\rho$:** rank correlation between $\hat{Q}_\phi(s,a)$ and a Monte-Carlo return proxy $\hat{G}(s,a)$, computed per state and averaged over states.
- **MAE:** mean absolute error $\mathbb{E}\left[\left|\hat{Q}_\phi(s,a) - \hat{G}(s,a)\right|\right]$ over sampled pairs.
- **ECE:** expected calibration error by binning $\hat{Q}_\phi$ values into quantiles and comparing bin means against empirical returns.
- **Advantage-energy baseline:** replacing $\hat{Q}_\phi$ with $\hat{A}_t$ in the energy weights to compare "advantage-based" vs "Q-based" guidance.

Empirically, $\hat{Q}_\phi$ provides a useful ranking signal on visited state–action pairs, and the combination of $\beta$-annealing, gating, and gradient clipping stabilizes guidance while the critic is still learning. These observations also support the bounded value-guidance error assumption used in the dual-proximal analysis.

### 5.8 Strict on-policy audit

We provide empirical evidence for strict on-policy learning using the three audit metrics defined in Section 4.5:

- **Offline Gradient Leakage Ratio (OGLR)** $\rightarrow 0$,
- **State Provenance Ratio (SPR)** $\rightarrow 1$,
- **Policy-Gradient Share (PGShare)** $\approx 1$.

Across tasks, OGLR remains at numerical-noise level, SPR stays close to 1.0, and PGShare typically lies in the 0.9–1.0 range. Together with the observation that **prior-KL is consistently much smaller than policy-KL**, these results indicate that: (i) policy gradients are dominated by the PPO surrogate computed on fresh rollouts, and (ii) synthetic proposals influence learning only through small-weight auxiliary branches without contaminating the PPO estimator.

### 5.9 Discussion: failure modes and practical boundary

Our results suggest that PPO-DAP is most effective when (i) the logged dataset provides at least **moderate coverage** of the relevant action support and (ii) the critic provides a **reasonably stable ranking signal** for value guidance.

Key limitations and failure modes include:

- **Dependence on logged-data coverage.** When $D_{\text{off}}$ has narrow or biased support, High-$Q$ hit rate and ESS decrease, and VG benefits diminish (Section 5.6). In extreme cases, PPO-DAP may effectively reduce to weak behavioral anchoring plus lightweight regularization, offering limited improvement over strong on-policy baselines.

- **Dependence on critic quality.** The effectiveness of VG is bounded by the ranking reliability and calibration of $\hat{Q}_{\phi}$. Although annealing/gating/clipping mitigate instability, sparse-reward tasks or non-stationarity can still make $Q$ −based guidance unreliable.
- **Scalability to higher-dimensional actions.** Proposal generation cost scales approximately linearly with $K$ and $N_{\text{steps}}$ , and overly aggressive proposal budgets may increase variance. In practice, moderate $K$, fast samplers, and moderate PET frequency provide a good balance.
- **Hyperparameter fairness vs. optimality.** We adopt aligned compute and shared tuning budgets across methods for fairness; more extensive per-method tuning could further improve absolute performance.

Overall, PPO-DAP provides a practical and auditable way to exploit logged data for faster on-policy learning: it improves early sample efficiency and often final returns **without changing PPO's estimator**, and it does so with **controllable engineering overhead** through explicit knobs ($K$, $N_{\text{steps}}$ , guidance strength, and PET frequency).

## 6. Conclusions and future work

This paper introduced PPO-DAP, a strictly on-policy framework that combines PPO with a value-guided diffusion action prior pretrained on logged trajectories. By decoupling decision making and generation and adapting the prior via parameter-efficient tuning, PPO-DAP improves exploration and sample efficiency while preserving an auditable on-policy update boundary.

We derived an informal dual-proximal performance lower bound that extends TRPO/PPO-style guarantees to the setting with a learned prior and value-guided proposals, highlighting the roles of policy KL, prior KL, and guidance error. Empirically, under a unified interaction budget of 1.0M environment steps across eight MuJoCo benchmarks, PPO-DAP consistently improves early learning efficiency and matches or exceeds strong on-policy baselines, while incurring only modest additional compute and memory overhead. Analysis of value guidance, coverage sensitivity, and strict on-policy audit metrics further clarifies when and why PPO-DAP is effective.

Limitations include the dependence on logged-data coverage, critic quality, and the cost of diffusion sampling in very high-dimensional action spaces. Future work includes extending PPO-DAP to multi-task or meta-RL settings, integrating model-based components, and applying the framework to real-world systems with richer observations and safety constraints.

## Appendix A. Notation and a proof sketch for the dual-proximal performance bound

### A.1 Main notation and quantities

For quick reference, we summarize the key notation used in Section 4.8 and in the dual-proximal bound (Eq. (14)).

- **On-policy actor**: $\pi_{\theta}(a \mid s)$. At online iteration $k$, we also write $\pi_k \triangleq \pi_{\theta^{(k)}}$.

- **Value head (critic)**: $V_\phi(s)$ trained by TD/GAE regression.
- **Lightweight action-value head (for VG only)**: $\hat{Q}_\phi(s,a)$. This head is used only to guide or rank candidate actions in the value-guided proposal (VG) module; it does **not** enter PPO's importance ratios or advantage estimation.
- **Diffusion action prior**: $p_\psi(a \mid s)$. During online learning, only a parameter-efficient subset ; $\psi_{\text{PET}} \subset \psi$ (e.g., Adapters/LoRA) is updated, while the diffusion backbone is frozen.
- **On-policy batch**: $D_{\text{on}}^{(k)}$, the rollout batch collected at iteration $k$ under $\pi_k$.
- **Synthetic proposal set**: $D_{\text{syn}}^{(k)}$ , the value-guided candidate actions generated from $p_\psi(a \mid s)$ on the *same* on-policy states $s \in D_{\text{on}}^{(k)}$.
- **PPO objective**: $\mathcal{L}_{\text{PPO}}$, the clipped surrogate with GAE advantages (Eq. (5)).
- **Actor objective**: $\mathcal{L}_{\text{actor}}$ , the total actor loss combining PPO with small-weight regularizers (prior KL and auxiliary imitation; Eq. (9)).
- **PET objective**: $\mathcal{L}_{\text{PET}}$ , the denoising loss used to adapt $\psi_{\text{PET}}$ online (the on-policy version of Eq. (6)).
- **Policy KL at iteration *k***:

$$\mathcal{K}_{\text{policy}}^{(k)} = \mathbb{E}_{s\sim\mathcal{B}^{(k)}}[\text{KL}(\pi_{k+1}(\cdot \mid s) \| \pi_k(\cdot \mid s))],$$

where $\mathcal{B}^{(k)}$ denotes the on-policy state batch induced by $D_{\text{on}}^{(k)}$.

• **Prior KL at iteration *k***: because diffusion models do not provide a convenient closed-form density, we use a diagonal-Gaussian proxy head

$$p_\psi^G(\cdot \mid s) = \mathcal{N}\left(\mu_\psi(s), \text{diag}\left(\sigma_\psi^2(s)\right)\right),$$

and define

$$\mathcal{K}_{\text{prior}}^{(k)} = \mathbb{E}_{s\sim\mathcal{B}^{(k)}}[\text{KL}(p_{k+1}^G(\cdot \mid s) \| p_k^G(\cdot \mid s))].$$

(Here $\text{KL}(\cdot \,\|\, \cdot)$ denotes the Kullback–Leibler divergence.)

• **Guidance-value error**: $\eta$ upper-bounds the discrepancy between the guidance estimator and the true action value under $\pi_k$, e.g.,

$$\eta \geq \sup_{(s,a)\in\mathcal{D}} \left|\hat{Q}_\phi(s,a) - Q^{\pi_k}(s,a)\right|$$

or is approximated empirically via one-step TD MAE or rank-quality measures (e.g., Spearman $\rho$) on visited $(s,a)$ pairs.

### A.2 Proof sketch for the performance lower bound

This section provides a **high-level** (informal) argument for Eq. (14), clarifying why the "dual-proximal" structure naturally yields a bound of the form "surrogate − policy-KL − prior-KL − guidance error. A fully rigorous derivation is outside the scope of this paper. Here we provide an informal sketch to clarify the intuition behind the dual-proximal bound.

**(1) Start from the performance difference lemma.** For any two policies $\pi_k$ and $\pi_{k+1}$, the

performance difference lemma states:

$$J(\pi_{k+1}) - J(\pi_k) = \frac{1}{1-\gamma} \mathbb{E}_{(s,a)\sim d_{\pi_{k+1}}} [A^{\pi_k}(s,a)]$$

where $d_{\pi_{k+1}}$ is the discounted visitation distribution induced by $\pi_{k+1}$.

The classical TRPO/PPO analysis constructs an **on-policy surrogate** by replacing $d_{\pi_{k+1}}$ with $d_{\pi_k}$ , and then bounds the resulting distribution mismatch error using a term proportional to $\mathrm{KL}(\pi_{k+1} \| \pi_k)$. This yields the familiar "surrogate minus policy-KL penalty" form.

**(2) View PPO-DAP as "a PPO step + small regularizers".** In PPO-DAP, the actor is optimized with

$$\mathcal{L}_{\text{actor}} = \mathcal{L}_{\text{PPO}} + \lambda_{\text{KL}} \mathcal{L}_{\text{prior}} + \lambda_{\text{aux}} \mathcal{L}_{\text{aux}} .$$

Crucially, both $\mathcal{L}_{\text{prior}}$ and $\mathcal{L}_{\text{aux}}$ are evaluated on the **same** on-policy states $s \in D_{\text{on}}^{(k)}$. Therefore, these terms do not introduce additional state-distribution shift beyond the standard PPO analysis; instead, they act as *soft constraints* on the policy outputs at already-visited on-policy states.

Intuitively, since the regularizers are small-weight perturbations of the PPO objective, their contribution to any degradation of the surrogate improvement can be controlled by (i) the induced drift of the prior (captured via $\mathcal{K}_{\text{prion}}^{(k)}$ ) and (ii) the accuracy of the value signal used to shape proposals (captured via $\eta$).

**(3) The prior-KL term arises from small-step PET updates.**

Online adaptation updates only $\psi_{\text{PET}}$ in a low-rank Adapter/LoRA subspace. Under a standard smoothness assumption (e.g., the denoiser is Lipschitz in parameters), a small parameter update $\Delta\psi_{\text{PET}}$ induces a small distributional change in the prior, and one can bound the drift as

$$\mathrm{KL}(p_{k+1} \| p_k) = \mathcal{O}(\|\Delta\psi_{\text{PET}}\|_2^2)$$

Hence, when PET uses sufficiently small steps, the prior drift is second-order small and can be controlled via the PET step size and update frequency, leading to a penalty term proportional to $\mathcal{K}_{\text{prior}}^{(k)}$ in the performance bound.

**(4) Aggregate value-guidance mismatch into $\boldsymbol{\eta}$.**

VG can be interpreted as reweighting or steering candidate actions using $\hat{Q}_\phi$. Any systematic discrepancy between $\hat{Q}_\phi$ and the true $Q^{\pi_k}$ introduces an additional approximation error in the improvement analysis. If this mismatch is bounded on visited state–action pairs, it can be summarized by a single penalty term $-c3\eta$. In experiments, the quality of $\hat{Q}_\phi$ can be assessed (and the assumption partially justified) via rank correlation, MAE, and calibration diagnostics.

**(5) Combine the terms to obtain Eq. (14).**

Putting the above components together—(i) a TRPO/PPO-style surrogate lower bound, (ii) a policy proximity penalty, (iii) a prior proximity penalty due to PET, and (iv) a guidance error

penalty—yields a bound of the form:

$$J(\pi_{k+1}) - J(\pi_k) \geq L_{\pi_k}(\pi_{k+1}) - c_1 \mathrm{KL}(\pi_{k+1} \| \pi_k) - c_2 \mathrm{KL}(p_{k+1} \| p_k) - c_3 \eta,$$

which matches Eq. (14) in the main text.

The key takeaway is that **dual proximity**—keeping both the policy update and the prior adaptation proximal—preserves the improvement behavior inherited from on-policy PPO, while explicitly accounting for the (monitorable) modeling error introduced by value-guided proposal selection.